\documentclass[11pt, a4paper]{article}
\usepackage[english]{babel}
\usepackage{amssymb,amsmath,amsthm,bm,graphicx}
\usepackage{authblk}
\usepackage[authoryear]{natbib}
\usepackage{multirow}
\usepackage{mathtools}
\usepackage[]{algorithm2e}
\usepackage[dvipsnames]{xcolor}
\usepackage{hyperref}

\newtheorem{theorem}{Theorem}

\theoremstyle{definition}

\newtheorem{remark}{Remark}

\newcommand{\indic}{{\bm 1}}
\newcommand{\BlackBox}{$\blacksquare$}

\def\beqn{\begin{eqnarray*}}
\def\eeqn{\end{eqnarray*}}
\def\beq{\begin{eqnarray}}
\def\eeq{\end{eqnarray}}
\def\bm#1{\mbox{\boldmath{$#1$}}}

\newcommand{\diag}{{\text{diag}}}

\begin{document}

\title{\Large\bfseries Dimension Estimation Using Random Connection Models}
\author{\bfseries Paulo Serra\thanks{
       This research took place while this author was a postdoctoral researcher at the Korteweg-de Vries Institute for Mathematics, of the University of Amsterdam, in Amsterdam, the Netherlands.}\, and Michel Mandjes\\ \vspace{-0.75em} {\rm Eindhoven University of Technology and University of Amsterdam}}
\date{\today}

\maketitle

\begin{abstract}
Information about intrinsic dimension is crucial to perform dimensionality reduction, compress information, design efficient algorithms, and do statistical adaptation.
In this paper we propose an estimator for the intrinsic dimension of a data set.
The estimator is based on binary neighbourhood information about the observations in the form of two adjacency matrices, and does not require any explicit distance information.
The underlying graph is modelled according to a subset of a specific random connection model, sometimes referred to as the Poisson blob model.
Computationally the estimator scales like $n\log n$, and we specify its asymptotic distribution and rate of convergence.
A simulation study on both real and simulated data shows that our approach compares favourably with some competing methods from the literature, including approaches that rely on distance information.
\end{abstract}


\section{Introduction}

In machine learning and computational geometry we often want to discover, or sometimes impose, structure on observations, and dimension plays a crucial role in this task.
The dimension of a data set can perhaps be best interpreted as the number of variables needed to describe it.
However, there is often a gap between the
ambient dimension of a data set -- the number of variables \emph{used} to describe it -- and its
intrinsic dimension -- the number of variables \emph{needed} to describe it (eventually up to a certain level of precision).
For instance,
high-dimensional data sets often live in lower dimensional spaces;
infinite-dimensional parameters of non-parametric models can often be accurately estimated using just a few parameters;
complex data can potentially be highly compressible.
However, this is usually not evident by just looking at the data, so in this paper we propose an algorithm to estimate the intrinsic dimension of a data set, and study its behaviour.

There are plenty of reasons to be interested in intrinsic dimensions.
Perhaps the most straightforward one is to perform dimensionality reduction.
Dimensionality reduction arises from a need to be able to extract meaningful conclusions from high-dimensional observations.
There is an extensive literature on this subject using multidimensional scaling, manifold learning, and projection techniques like principal component analysis and projection pursuit;
for a general overview cf.~\citep{fodor2002survey, burges2010dimension, lee2007nonlinear}.
For specific techniques see for example~\citep{kohonen1990self, cox2000multidimensional, tenenbaum2000global, roweis2000nonlinear, donoho2003hessian, huo2002local, gine2006empirical}, and the references therein.
In order to be used to their full extent, these approaches require a priori knowledge about the intrinsic dimension of the data set.
Knowledge about intrinsic dimension is also important in independent component analysis; cf.~\citep{hyviirinen2001independent}.

From a statistical perspective, the intrinsic dimension provides information about the difficulty of making inference.
Non-parametric estimators usually rely on approximation properties of certain function spaces;
the dimension of the support of these functions influences these approximation properties.
Statistical adaptation often focuses on smoothness, but dimension actually has a more substantial impact on rates of convergence.
Knowledge about dimension is also important to avoid (if possible) the curse of dimensionality.
Dimension plays an important role in classification problems as well, where performance is greatly compromised in high dimensions;
cf.~\citep{bickel2004some, fan2008high}.
There are also connections to search, and to outlier detection; cf.~\citep{amsaleg2015estimating} and the references therein.

From a computational perspective, the dimension of a data set impacts the amount of space needed to store data (compressibility).
The speed of algorithms is also commonly affected by the dimension of input data.
Learning the underlying dimension is also important to design algorithms that require less data (meaning, make better use of available data) when data happen to live on a low dimensional space.
Because of this, knowledge about dimension is crucial in many fields such as biomedicine, economics, engineering, astronomy, remote sensing, and computer vision, with important applications in mass spectrometry, genetics, networking, image processing, automatic text analysis, among others;
for some concrete applications see~\citep{verleysen1999forecasting, lahdesmaki2005intrinsic, abrahao2008internet, carter2010local} and references therein.

Early work on dimension estimation dates back to~\citep{shepard1962analysis1, shepard1962analysis2, kruskal1964multidimensional, kruskal1964nonmetric, bennett1969intrinsic} on multidimensional scaling.
The idea is that one has measurements of similarities (or dissimilarities) between data points, and would like to find points in a potentially high-dimensional vector space that are consistent with the observed similarities (resp.\ dissimilarities).
Other approaches such as that of~\citep{fukunaga1971algorithm, fukunaga198215} are based on Karhunen--Lo\`eve expansions.
There, and much in the spirit of principal component analysis, the idea is that minimal representations that capture most of the variation in the data carry information about dimensionality; see also an approach based on testing by~\citep{trunk1968statistical}.
Information criteria such as AIC by~\citep{akaike1974new} and BIC by~\citep{schwarz1978estimating} can also be used to estimate dimensions within a model.

More recent work of~\citep{camastra2002estimating, kegl2002intrinsic, hein2005intrinsic, sricharan2010optimized} is based on the notion of correlation integral; cf.~\citep{grassberger2004measuring}.
The correlation dimension is a limit associated with this correlation integral.
This notion of dimension, which we also use in this paper, is computationally attractive when compared with other notions of dimension such as that of box counting dimension, for example.
There are also the techniques of~\citep{costa2004learning, farahmand2007manifold, leonenko2008class, kleindessner2015dimensionality} based on different types of graphs where edges represent some form of proximity, like k-nearest neighbour graphs, and geodesic minimal spanning trees.
Their work relies on the fact that certain quantities associated with these graphs scale monotonically with dimension.
By exploring this, one can extract information about dimensionality by inverting these relations.
A similar idea is used in~\citep{amsaleg2015estimating} by recurring to notions from extreme value theory.
See also~\citep{eriksson2012estimating} for a clustering-based approach, and the work of~\citep{levina2004maximum} for an estimator based on an approximation of the likelihood.

There is some room for improvement in the approaches mentioned above.
Some of them rely on rather extensive knowledge about distances between all possible pairs of observations, sometimes also of perturbations thereof, or on certain hierarchical constructs like dendrograms;
because execution times tend to scale quadratically with the sample size and linearly with dimension, these quickly become computationally costly as the number of observations or the dimension of the measurements is high -- exactly the situation where dimension estimation is most important.
(In genetics and computer vision applications, for instance, the number of observed dimensions can easily reach hundreds of thousands.)
Therefore, either due to the volume or to complexity of the data, we may be computationally limited to work with very crude information, such as knowing only whether each pair of observations is close or not.
Because of this, it is of interest to develop methods to estimate dimension that rely on as little information as possible.

Another aspect that is often overlooked in the literature is that the intrinsic dimension of a data set is usually scale-dependent: the dimension of the data set depends on the scale at which we analyse it; cf.~\citep{burges2010dimension}.
Say we sample points uniformly at random on a manifold with noise; if we look at the data set on a fine scale we only pick up on the noise, whereas at a larger scale the features of the manifold will dominate.
The manifold itself can have different dimensions depending on which scale we look at it, and the noise may have arbitrary dimension.
It is therefore not clear what ``the dimension of a data set'' is, unless we specify a scale to go with it (or if the support of the distribution of the data is homogeneous or unstructured).
However, even then the dimension is very dependent on the specific data (e.g., structure of the manifold, distribution of the noise).
Approaches based on regressing the logarithm of the correlation integral on the logarithm of its argument fail to capture this and instead return something akin to an average dimension across scales for the data set.
Approaches based on k-nearest-neighbour graphs also have limitations with regards to this;
the distance from a datapoint to its k-nearest neighbour scales in a non-trivial way with dimension and is quite dependent on the distribution of the observations.
This makes it difficult to estimate dimension by inversion without specific knowledge on the distribution of the data.\par

In this paper we resolve the limitations identified above.
We estimate the intrinsic dimension of a data set at a user-prescribed scale based solely on binary neighbourhood relations between observations.
More specifically, we assume that certain undirected graphs (or their adjacency matrices) can be observed.
In this graph, each vertex corresponds to an observation that lives in some high- (possibly infinite-) dimensional space.
An edge is present between two vertices if the corresponding observations are \emph{close}.
What we consider to be close determines the scale at which we analyse the data.
The goal is to estimate the intrinsic dimension of the data set based on the adjacency matrix of the graph only, i.e., without explicit access to distance information.
We model such a graph according to a subset of a random connection model, a model from continuum percolation; cf.~\citep{penrose1991continuum}, and~\citep{meester1996continuum} for an overview on the subject.
More specifically, we model it according to a subgraph of a graph sampled from the so called Poisson blob model; cf.~\citep{grimmett1989percolation}.

We propose an estimator based on the doubling property of the Lebesgue measure and on the notion of correlation integral.
The estimator does not rely on distance information about the observations and has computation time that scales like $n\log n$, which is particularly important when dealing with large, high-dimensional data sets.
Since only (sparse) adjacency matrices have to be stored, our approach also leads to a reduction of the required storage space.
Under an identifiability condition, we show that the estimator is consistent and asymptotically Gaussian, and we compute its rate of convergence.
To the best of our knowledge such results are not yet available in the literature.
The estimator strongly concentrates around its expectation but in general the constants involved in the rate scale exponentially with the intrinsic dimension;
the bias plays an important role as it is the main bottleneck in the procedure leading to a logarithmic rate.
We propose a bias corrected estimator that follows the same (optimal) asymptotics, but which performs much better according to our numerical experiments.

Minimax rates are unknown for the type of data that we consider, but in the (easier) case where one has access to the actual observations, these can be found in~\citep{koltchinskii2000empirical} and are logarithmic.
(For the noiseless case see~\citealp{wasserman:2016}.)
This means that our procedure is essentially optimal and that its computational efficacy is not obtained at the expense of precision.
Furthermore, we are capable of producing estimates of the spread of the estimator, without a need for resampling, and these quantify the uncertainty in the estimate fairly well.
This is particularly important given the  slow convergence rate, and is a major improvement over competing approaches, which tend to overly concentrate around biased estimates.
We also run some numerical simulations that show that our estimator compares favourably with competing estimators (particularly when it comes to recuperate an integer dimension), including estimators that rely on distance information.\par

This paper is structured as follows.
In Section~\ref{sec:model} we formally define our model.
The interpretation of the effect of scale in the model is given in Section~\ref{sec:role_of_epsilon}.
Section~\ref{sec:estimates_d} has a description of our estimator.
Section~\ref{sec:estimates_p1} contains consistency results for a relevant probability in the model.
Section~\ref{sec:consistency_d_epsilon} has our main result about the consistency of our estimator for the intrinsic dimension, and a comparison with related work from the literature.
In Section~\ref{sec:simulations} we present some numerical illustrations for our method, and we propose our bias corrected estimator.
We close with some conclusions in Section~\ref{sec:discussion}.
The proofs of our main results are collected in the Appendix.

\section{Sampling, Model, Notation, and Problem Formulation}\label{sec:model}

Consider the following model.
Sample \emph{design points} $X_1,\dots,X_n\in\mathbb{R}^D$, independently, from a distribution $F$, where $D\in\mathbb{N}$ is some \emph{ambient dimension}.
Given the design points $\bm X = (X_1,\dots,X_n)^T$, construct a random undirected graph by placing an edge between two vertices $i<j\in\{1,\dots, n\}$ if $r(X_i,X_j)\le\epsilon$, $\epsilon>0$, where $r$ is a metric\footnote{All of the assumptions on the metric will be implicit.} on $\mathbb{R}^D$.
We denote the adjacency matrix of the resulting random graph as $\bm A_\epsilon$.
If we disregard that our design points are typically concentrated (since they are sampled from $F$), this is a subset of a graph sampled from what is usually called the Poisson blob model, a model from continuum percolation.
In this model we allow $\epsilon=\epsilon_n$ to converge to $0$ as $n\to\infty$, if need be;
we discuss the role of $\epsilon$ in more detail in Section~\ref{sec:role_of_epsilon}.
Most of the quantities that we define in the following depend on $n$ but this is omitted from the notation except when $n$ plays a role.

We assume that the observations in our data set actually live (potentially in approximate form) in a lower (potentially fractional) dimensional space.
For instance, the design points can have the form $X_i = \varphi(\tilde X_i) + \sigma\cdot\epsilon_i$, $\sigma\ge 0$, where $\varphi:\mathbb{R}^d\mapsto\mathbb{R}^D$, $d\le D$, is some embedding.
The observations can therefore be highly structured; they can be concentrated around, say, a manifold.
The number $d$ is called the \emph{intrinsic dimension} of the data set $\bm X$, and it is our object of interest.

Our statistical problem is the following:
for a data set with $n$ observations we have access to a symmetric, binary matrix $\bm{\mathcal A}$ where $\mathcal{A}_{i, j}=\mathcal{A}_{j, i}=1$ if, and only if, the $i$-th and $j$-th observations are ``close'';
the data points (or distances between data points) are not actually observed.
We assume that our notion of ``close'' is reasonable, in the sense that we can model $\bm{\mathcal A}$ according to a random connection model:
 $\bm{\mathcal A} = \bm A_\epsilon$ for some $\epsilon$ and some metric $r$ (which are not necessarily known to us).
Given access to such adjacency matrices\footnote{
In fact we consider two adjacency matrices as it will become clear from the definition of our estimator in~\eqref{def:estimators_d_implicit}.
We argue in Section~\ref{sec:estimates_d}, that this is inevitable since $\epsilon$ trades off with the standard deviation of $F$.
} we would like to estimate the intrinsic dimension $d$.
The point is that although the support of $F$ may be high-dimensional, most of the mass of $F$ might be concentrated on a (lower) dimensional sub-space or manifold (eventually as $n\to\infty$, or $\sigma\to 0$), such that one can find a parsimonious representation for that data that still preserves its main features.

We denote by $\bm B_\epsilon$ the degrees of the vertices in the graph such that $\bm B_\epsilon = (B_{\epsilon,1}, \dots, B_{\epsilon, n})$,
\begin{equation}\label{def:B_entries}
B_{\epsilon, i}
= \sum_{j=1}^n A_{\epsilon, i, j},
 \quad i=1,\dots, n,
\end{equation}
where the (binary) entries $A_{\epsilon, i, j}$ of the adjacency matrix $\bm A_\epsilon$ satisfy
\begin{equation}\label{def:A_entries}
A_{\epsilon, i, j} = A_{\epsilon, j, i} = \indic_{\{r(X_i,X_j)\le \epsilon\}}, \qquad\hbox{and}\qquad
A_{\epsilon, i, i} = 0, \quad i, j=1,\dots, n, \, i\neq j.
\end{equation}
By construction, the distribution of the random matrix $\bm A_\epsilon$ is invariant under any permutation of its rows and columns so that the $B_{\epsilon, i}$ are identically distributed but not independent.
Define the two functions
\begin{equation}\label{def:connection_probability}
p_\epsilon(x) = \mathbb{P}\{r(X, x)\le\epsilon\},
\qquad\hbox{and}\qquad
p_\epsilon(x, y) = \mathbb{P}\{r(X, x)\le\epsilon, r(X, y)\le\epsilon\},
\end{equation}
where the probability is taken with respect to $X\sim F$.
This is the (local) connection probability for a design point at site $x$, and the probability of two design points at sites $x$ and $y$ sharing a neighbour.
With this notation,
\begin{equation}\label{eq:dist_B}
B_{\epsilon, i}|X_i \sim \hbox{Bin}\{n-1,\, p_\epsilon(X_i)\}. %,
\end{equation}
From this we see that if $p$ were constant, then the model would reduce to the Erd\H{o}s--R\'enyi model of~\citep{erdHos1959random}.
If $p_\epsilon$ (which depends exclusively on $F$ and $\epsilon$) is not constant, then this leads to some inhomogeneity for the resulting random graph.

In what follows we denote, for $i, j, k$ mutually different,
\begin{equation}\label{def:p1_p2}
p_{\epsilon,1} = \mathbb{E}A_{\epsilon, i, j},
\qquad\hbox{and}\qquad
p_{\epsilon,2} = \mathbb{E}A_{\epsilon, i, k}A_{\epsilon, k, j}.
\end{equation}
These two numbers (or sequences, if $\epsilon\to0$) are the probability that two vertices connect and the probability that two vertices have a common neighbour, respectively.
Note that if $X,Y,Z$ are independent and distributed according to $F$, then
\[
p_{\epsilon,1} =
\mathbb{P}\{r(X,Y)\le\epsilon\} =
\mathbb{E}\mathbb{P}\{r(X,Y)\le\epsilon|X\} =
\mathbb{E}p_\epsilon(X),
\]
and in the same way,
\[
p_{\epsilon,2} =
\mathbb{E}\mathbb{P}\{r(X,Z)\le\epsilon, r(Z,Y)\le\epsilon|X,Y\} =
\mathbb{E}p_\epsilon(X,Y).
\]
By definition, $p_\epsilon(X,X)=p_\epsilon(X)$.
Also, by independence and Jensen's inequality,
\[
p_{\epsilon,2} =
\mathbb{E}\mathbb{P}\{r(X,Z)\le\epsilon, r(Z,Y)\le\epsilon|Z\} = \mathbb{E}\{p_\epsilon(Z)^2\} \ge \mathbb{E}\{p_\epsilon(Z)\}^2 = p_{\epsilon,1}^2.
\]
In fact, the (non-negative) difference $p_{\epsilon,2}-p_{\epsilon,1}^2$ is the variance of the connection probability function $p_\epsilon(x)$ which will play an important role later in the paper.

Both $p_{\epsilon,1}$ and $p_{\epsilon,2}$ depend on $\epsilon$ (and $F$), but also on the dimensionality of the data.
For example, it is clear that $p_{\epsilon,1}$ and $p_{\epsilon,2}$ decrease as $\epsilon$ decreases.
In fact, most of what follows is based on this dependence.
Before we give the intuition behind our estimator, we discuss the role of $\epsilon$ in our approach.

\section{Role of $\epsilon$ in the Model}\label{sec:role_of_epsilon}

\begin{figure}[!tb]
\centering
\includegraphics[trim={6cm 9cm 6cm 9cm}, width=0.3\textwidth]{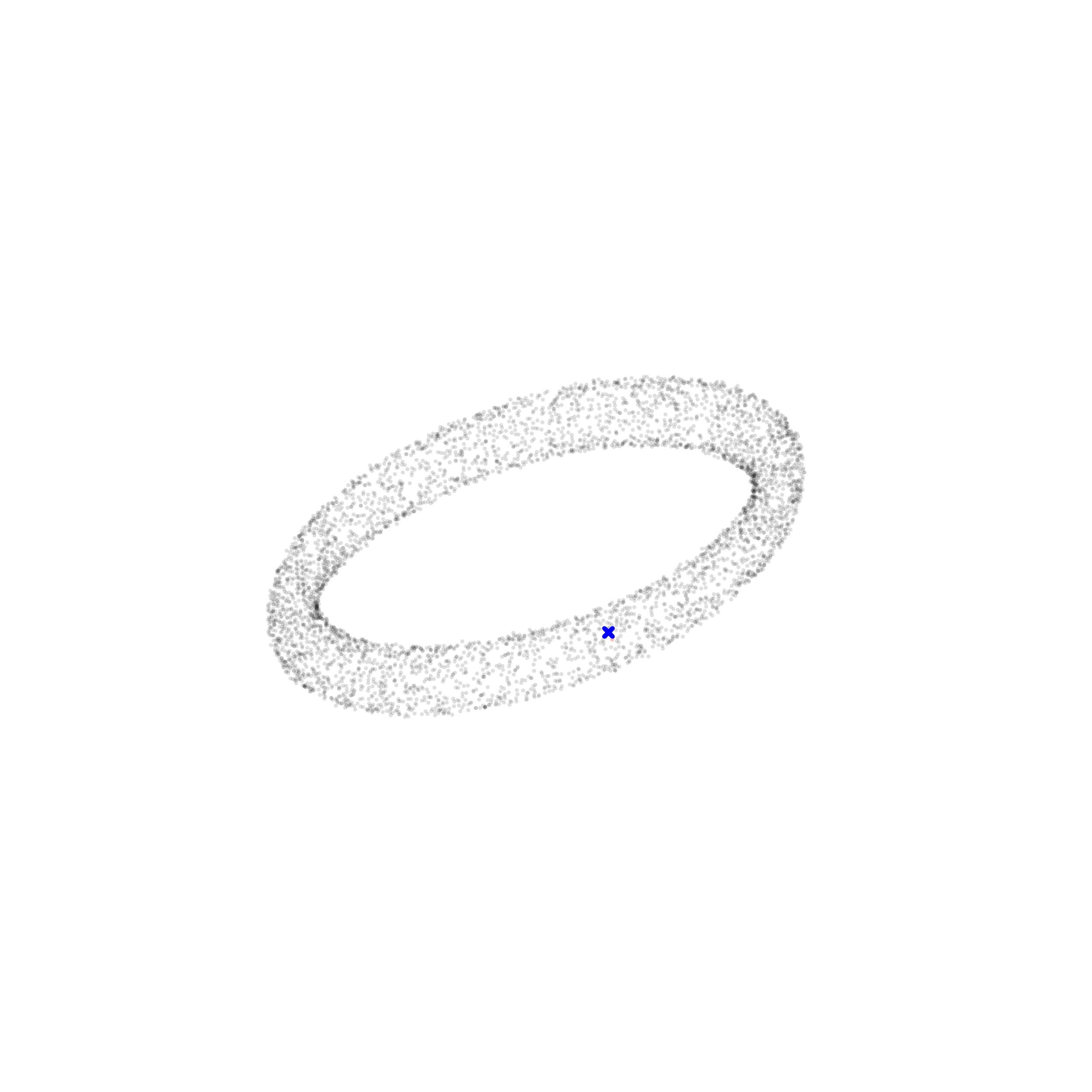}\includegraphics[trim={6cm 9cm 6cm 9cm}, width=0.3\textwidth]{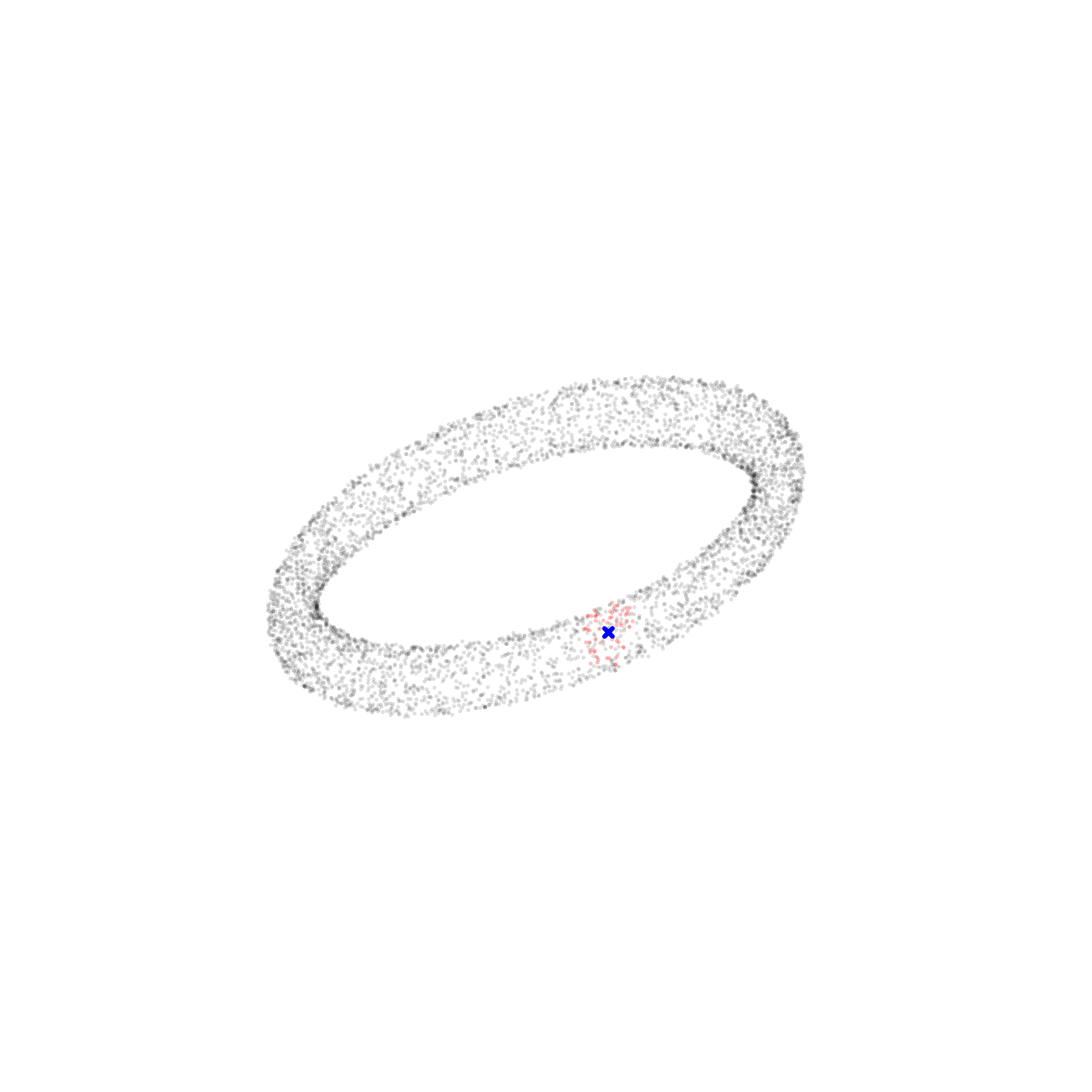}\includegraphics[trim={6cm 9cm 6cm 9cm}, width=0.3\textwidth]{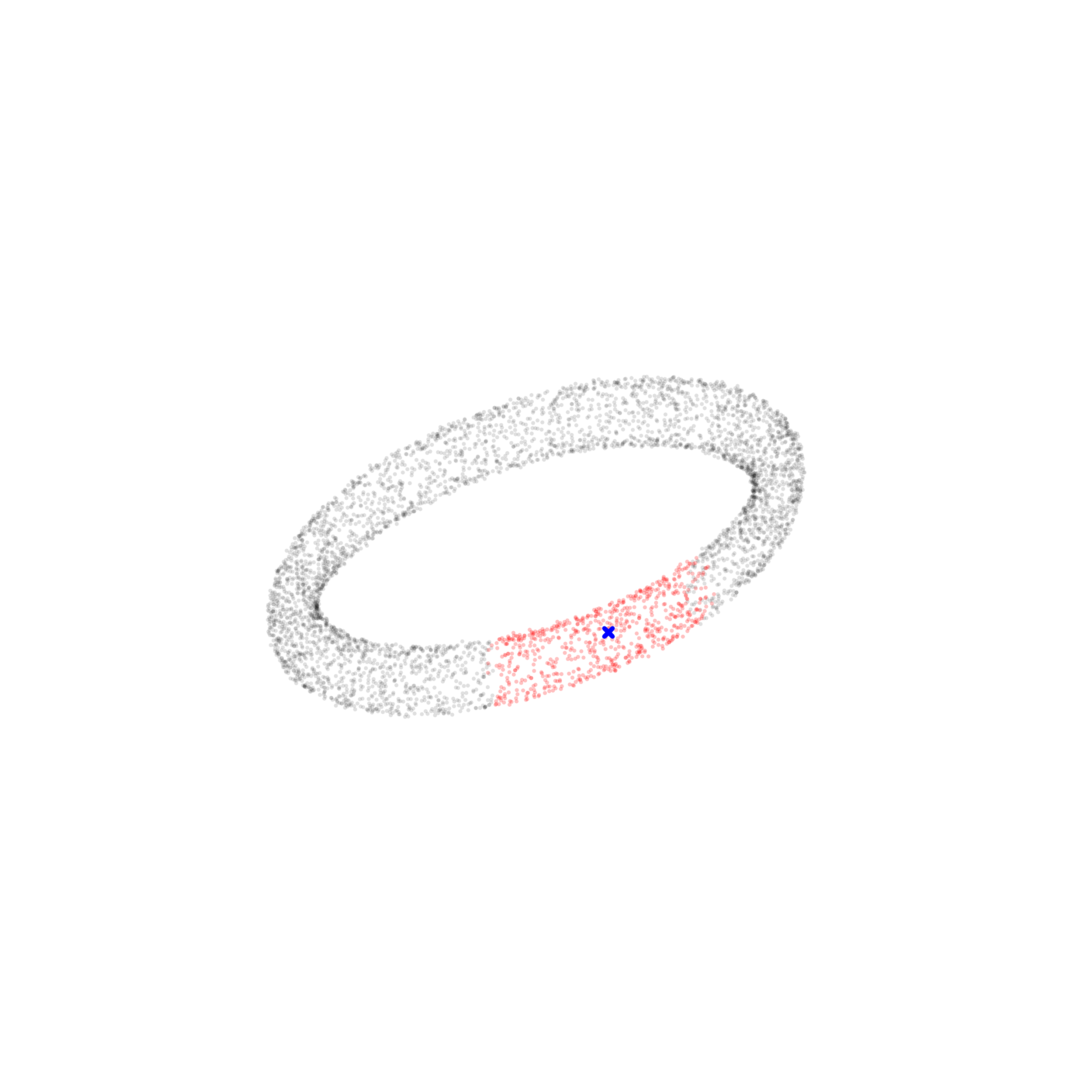}
\caption{\footnotesize
Design points sampled uniformly on a torus, with noise.
Design points within an $\epsilon$ distance (increasing from left to right) of a fixed design point are coloured red.
}\label{fig:intersections_manifold}
\end{figure}

The parameter $1/\epsilon$ can be seen as a resolution level that determines at which distance we distinguish between design points.
This parameter plays a crucial role in our approach as is explained in this section.
In Figure~\ref{fig:intersections_manifold} we exemplify the effect of the size of $\epsilon$.
We sampled points uniformly at random on a manifold, then added some (3-dimensional) Gaussian noise;
these points are the design points $\bm X$ and are embedded on a 3-dimensional space.
We then took one of the design points, and coloured red all points that fall within a given Euclidean distance $\epsilon$ of the selected design point;
the three plots correspond to different choices of $\epsilon$.

If $\epsilon$ is so small that no red dots would be present, then the detected dimension is $0$.
If $\epsilon$ is large enough to capture just a few nearest neighbours (leftmost plot), then we capture only the effect of the noise -- 3-dimensional in our case -- but arbitrary in general.
Increasing $\epsilon$ (central plot), the intersection is now a 2-dimensional section of the surface of the manifold.
Further increasing $\epsilon$ (rightmost plot) changes the shape of the intersection which is now the (1-dimensional) surface of a tube.
In either case, with much larger choices for $\epsilon$ we would capture all design points and the volume of the intersection would vanish with respect to $\epsilon$; this would again lead the trivial case where dimension $0$ is detected.

Of course the dimension can also be fractal, and we could also be interested in the dimension of just a region of the manifold in which case the choice of $\epsilon$ (and the design point that defines the neighbourhood) plays an important role again.
The bottom line is that $\epsilon$ should be considered a parameter of the model (as opposed to a parameter of the estimator) in that the target intrinsic dimension should be seen as a function of $\epsilon$; cf.~\citep{burges2010dimension} for a similar discussion.
In other words, the resolution level should be chosen in line with the goals of the analysis;
to learn the structure of the noise one would pick relatively small values of $\epsilon$, while to learn the dimension of the manifold we would have to pick larger values.
This is related to the scale of the observations, and it should be taken into consideration when choosing $\epsilon$.
Another issue is that one should also account for the sample size in the form of a finite sample correction.
We return to this point in Section~\ref{sec:simulations} where we use some numerical experiments to illustrate this point.

Next we define and give the intuition for our estimator of the intrinsic dimension $d$.

\section{Estimation of the Intrinsic Dimension}\label{sec:estimates_d}

We start by providing some heuristic motivation for our estimator of $d$.
Consider, for $x\in\mathbb{R}^m$, $m\in\mathbb{N}$ the balls $V_\epsilon(x,m) = \{y\in\mathbb{R}^m: r(x, y)\le\epsilon\}$ for a homogeneous, translation-invariant metric $r$, and denote $V_\epsilon(m) = V_\epsilon(0,m)$.
Assume, without loss of generality, that $0\in\mathcal{X}\subseteq\mathbb{R}^D$, where $\mathcal{X}$ is an appropriate high probability set in the support of $F$.
If $\epsilon$ is small (or if $\epsilon\to0$) and if $F$ admits a continuous density $f$ with respect to the Lebesgue measure $\mu$, then we expect
\[
p_\epsilon(x) \approx
\int_{\mathcal{X}} \indic_{V_\epsilon(x,D)}(y)\,f(y)\, d\mu(y) \approx
f(x)\cdot \int_{\mathcal{X}} \indic_{V_\epsilon(D)}(y)\, d\mu(y) \triangleq f(x)\cdot v_\epsilon.
\]
The assumption that the intrinsic dimension of the data set is $d$, corresponds to assuming, with mild abuse of the notation, that (for all appropriately small $\epsilon$)
\[
v_\epsilon \triangleq
\mu\{V_\epsilon(D)\cap\mathcal{X}\} \approx
\mu\{V_\epsilon(d)\}.
\]
(In fact, this relation can be seen as a definition of the $\epsilon$-scale intrinsic dimension.)
This has the interpretation that at the $\epsilon$-scale the data looks $d$-dimensional.
The point is that $p_\epsilon(x)$ should not be sensitive to the dimension $D$ of the data points, but instead to the intrinsic dimension $d$ of the data set and an $\epsilon$-scale. %(and to $\epsilon$).

Since $p_{\epsilon,1} = \mathbb{E}\{p_\epsilon(X)\}$, we can approximate
\[
p_{\epsilon,1} \approx
\mathbb{E}f(X)\cdot \mu\{V_\epsilon(d)\}.
\]
One can estimate $d$ by replacing $p_{\epsilon,1}$ by an estimator and inverting the relation above.
However, this would only be feasible with knowledge of the distribution $F$ via the constant $\mathbb{E}f(X)$ and of the parameter $\epsilon$, which in general we do not have access to.

Arguably, the most reasonable way to get rid of the dependence on $F$ and $\epsilon$ is to examine the data at two different scales simultaneously.
For appropriately small $\epsilon$,
\begin{equation}\label{eq:heuristic_approximations}
\frac{p_{2\epsilon,1}}{p_{\epsilon,1}} \approx \frac{
\mu\{V_{2\epsilon}(d)\}
}{
\mu\{V_\epsilon(d)\}
}
\cdot
\end{equation}
This is a natural idea.
Looking back at Figure~\ref{fig:intersections_manifold}, the (hyper-)volume of the intersections (which can be inferred from the number of points in the intersection) does not give us any information about dimension;
it is how this quantity \emph{scales} with $\epsilon$ that is informative.

With this approximation in mind, we define an estimator for $d$:
for an arbitrary function $g_\epsilon(d)$ -- ideally $p_{2\epsilon,1}/p_{\epsilon,1}$, but in general any reasonable approximation of $\mu\{V_{2\epsilon}(d)\}/\mu\{V_{\epsilon}(d)\}$ -- the estimator is defined as (any) implicit solution $\hat d_n$ to the equality
\begin{equation}\label{def:estimators_d_implicit}
\frac{\hat p_{n,2\epsilon,1}}{\hat p_{n,\epsilon,1}} = g_\epsilon\big(\hat d_n\,\big),
\end{equation}
where $\hat p_{n,\epsilon,1}$ is any estimator for $p_{\epsilon,1}$, $\epsilon>0$.
If $d$ is an integer, then we can consider an estimator $[\hat d_n\,]$, where $[\,\cdot\,]$ represents the argument rounded to the closest integer.
Note that the function $g_\epsilon$ is allowed to depend on $n$.

The need to look at the data at two different scales simultaneously should not be a surprise. The probability $p_{\epsilon,1}$ itself does not carry any information about dimension if $F$ is unknown;
it is instead how $p_{\epsilon,1}$ scales as a function of $\epsilon$ that provides information about dimension.
This notion of scaling is in fact connected with the notion of expansion dimension of~\citep{Karger:2002:FNN:509907.510013}.

For a given metric $r$ one can numerically approximate the function $\mu\{V_{2\epsilon}(d)\}/\mu\{V_{\epsilon}(d)\}$, but in analogy to the doubling property of the Lebesgue measure this function should be constant over $\epsilon$, at least if $\epsilon$ is appropriately small.
So a canonical choice (independent of $\epsilon$) would be $g_\epsilon(d) = 2^d$, in which case one has an explicit estimator for $d$:
\begin{equation}\label{def:estimators_d_explicit}
\hat d_n = \frac{\log \hat p_{n,2\epsilon,1} - \log \hat p_{n,\epsilon,1}}{\log 2}\cdot
\end{equation}
This is just an example of a possible form that the estimator can take.
However, it does suggest that at least for certain models one can expect to have explicit estimators for $d$ that do not require knowledge of $\epsilon$, $F$, or $r$ and are therefore completely parameter-free.

Although $\epsilon$ should be known (or indeed picked, as pointed out in the discussion in Section~\ref{sec:role_of_epsilon}), one can define reasonable estimators for $d$ in the case where $\epsilon$ is unknown.
In fact, it is is not possible to estimate $\epsilon$ consistently from adjacency matrices $\bm A_\epsilon$ without knowledge of the distribution $F$.
This follows from the trade-off between the standard deviation of the distribution $F$ and the radius $\epsilon$.
Let $\sigma>0$, and say $\bm A_{\epsilon/\sigma}$ is associated with design points $X\sim F$ and $\bm A_{\epsilon}'$ is associated with design points $X'\sim F(\cdot/\sigma)$;
then
\[
A_{\epsilon/\sigma, i, j} =
\indic_{\{r(X_i,X_j)\le \epsilon/\sigma\}} =
\indic_{\{r(\sigma X_i,\sigma X_j)\le \epsilon\}} \sim
A_{\epsilon, i, j}',
\]
so that information about $\epsilon$ cannot be retrieved from the adjacency matrix without knowledge of the distribution $F$.
However, if $\epsilon$ is known, one may try to flesh out lower order terms in the approximation above to reduce the bias of the estimates.

We can more precisely approximate the local connection probability as
\[
p_\epsilon(x) \approx
f(x)\cdot v_\epsilon + \int_{\mathcal{X}} \indic_{V_\epsilon(x,D)}(y)\,(y-x)^T\nabla f(x)\, d\mu(y),
\]
where $\nabla f$ denotes the gradient of the density $f$, so that by taking expectation,
\[
p_{\epsilon,1} \approx
\mathbb{E}f(X)\cdot v_\epsilon\cdot \big( 1+ \Delta_\epsilon \big), \qquad
\Delta_\epsilon = \frac{\mathbb{E}\int_{\mathcal{X}} \indic_{V_\epsilon(X,D)}(y)\,(y-X)^T\nabla f(X)\, d\mu(y),}{\mathbb{E}f(X)\cdot v_\epsilon}\cdot
\]
For our canonical choice $g_\epsilon(d)= 2^d$ we thus obtain the approximation
\[
\frac{\log p_{2\epsilon,1} - \log p_{\epsilon,1}}{\log 2} \approx
\frac{\log v_{2\epsilon} - \log v_\epsilon + \log\big(1+\Delta_{2\epsilon}\big) - \log\big(1+\Delta_\epsilon\big)}{\log 2} \approx
d + \frac{\Delta_{2\epsilon} - \Delta_\epsilon}{\log 2}.
\]
By the Cauchy-Schwarz inequality to the inner product $(y-x)^T\nabla f(x)$, we conclude that
\[
\frac{|\Delta_{2\epsilon} - \Delta_\epsilon|}{\log 2} \le \frac3{\log 2}\cdot\frac{\mathbb{E}\|\nabla f(X)\|}{\mathbb{E} f(X)} \cdot\epsilon.
\]
Although the multiplier above is unknown, it depends only on $F$ and is therefore fixed.
Furthermore, it is reasonable to expect the multiplier to be of order $d$, since the gradient of the density should only be non-trivial along $d$ independent directions.
This means that certain choices for the function $g_\epsilon(d)$, like for example choices that are independent of the scale $\epsilon$, should result in a bias of order $O(d\cdot\epsilon)$.
We return to this discussion in Section~\ref{sec:consistency_d_epsilon} after we have specified the asymptotics of $\hat d_n$ for arbitrary $g_\epsilon$.

\begin{remark}\label{rem:p2vsp1}
A similar reasoning to the one that was applied to $p_{\epsilon,1}$ above can be applied to other probabilities associated with the model, like for example $p_{\epsilon,2}$, to motivate alternative estimators for the intrinsic dimension.
Although not reported here, we did not find any noticeable difference between the $\hat d_n$ estimator defined in~\eqref{def:estimators_d_implicit} and a $p_{\epsilon,2}$ based estimator.
\end{remark}

From the discussion above, it is clear that the consistency of the estimators defined in~\eqref{def:estimators_d_implicit} depends on three factors:
the consistency of the estimates of $p_{\epsilon,1}$ that are used, the quality of the approximation in~\eqref{eq:heuristic_approximations}, and the slope of $g_\epsilon(d)$ around the underlying intrinsic dimension $d$.
First we address the estimation of $p_{\epsilon,1}$.

\section{Estimates of the Connection Probability and their Asymptotics}\label{sec:estimates_p1}

An estimator for $p_{\epsilon,1}$
 is obtained by averaging off-diagonal entries of the matrix $\bm A_\epsilon$.
For any $m_n\le n$ ($m_n$ is for now left unspecified, and is a parameter of the estimator),
\begin{equation}\label{def:p1_hat}
\hat p_{n,\epsilon,1} =
\frac1{m_n} \sum_{i=1}^{m_n}\frac{B_{\epsilon, i}}{n-1} =
\frac2{m_n(n-1)} \sum_{i=1}^{m_n}\sum_{j=i+1}^nA_{\epsilon, i, j},
\end{equation}
using the symmetry of $\bm A_\epsilon$.
Since $\mathbb{E}B_{\epsilon, i}/(n-1)=p_{\epsilon,1}$, the estimator $\hat p_{n,\epsilon,1}$ is unbiased.
This estimator can be evaluated in $O(n\,m_n)$ instructions.
If we set $m_n=n$ then the execution time may be prohibitive if $n$ is large so the parameter $m_n$ offers some flexibility.
However, as we will see below and in Sections~\ref{sec:consistency_d_epsilon} and~\ref{sec:simulations}, the role of the sequence $m_n$ goes beyond just controlling the computational complexity of the estimator.
In Section~\ref{sec:simulations:choice_m_n}, in particular, we discuss what constitutes a ``good'' choice for $m_n$.

The following theorem provides the asymptotics for the estimator in~\eqref{def:p1_hat}.

\begin{theorem}\label{theo:rate_p1}
Let $m_n\le n$ and $m_n\to\infty$ as $n\to\infty$.
If $m_n=o(n)$, and $p_{\epsilon,2}>p_{\epsilon,1}^2$, then
\begin{equation}\label{eq:asymptotics_p1}
S_{n,\epsilon,1}^{-1/2} \cdot \left\{\frac{\hat p_{n,\epsilon,1}}{p_{\epsilon,1}}-1\right\} \stackrel{d}{\longrightarrow} N(0,1), \qquad \hbox{where}\qquad
S_{n,\epsilon,1} = \frac{p_{\epsilon,2}-p_{\epsilon,1}^2}{m_n\, p_{\epsilon,1}^2}.
\end{equation}
If $m_n=n$ then the previous display also holds if we further assume that %$p_{\epsilon,1}-p_{\epsilon,2}=o\{n(p_{\epsilon,2}-p_{\epsilon,1}^2)\}$, and that
$n(p_{\epsilon,2}-p_{\epsilon,1}^2)^2\to\infty$.
(This assumption always holds if $\epsilon$ is fixed.)
\end{theorem} \hfill \BlackBox

The proof of this theorem can be found in the Appendix.
This result is valid irrespectively of the distribution $F$, and holds even if $\epsilon\to0$, as $n\to\infty$, so that $p_{\epsilon,1},p_{\epsilon,2}\to0$.
The difference $p_{\epsilon,2}-p_{\epsilon,1}^2 = \mathbb{E}\{p_\epsilon(X)^2\}-\{\mathbb{E}p_\epsilon(X)\}^2$ is the variance of the function $p_\epsilon$.
From this we see that $p_\epsilon$ being more variable has a negative impact on the estimation of $p_{\epsilon,1}$, which is not surprising.
If $\epsilon$ is fixed, then $p_{\epsilon,1}$ can be estimated with rate $m_n^{-1/2}$.
However, if $\epsilon\to0$ the rates may be different depending on how the probabilities involved scale with $\epsilon$, which in turn depends on the specific distribution $F$ and the metric $r$ at hand.

Next we give conditions under which the estimators~\eqref{def:estimators_d_implicit} are consistent for $d$.

\section{Consistency of Estimates for the Intrinsic Dimension}\label{sec:consistency_d_epsilon}

Based on the asymptotics of $\hat p_{n,\epsilon,1}$ %and $p_{\epsilon,2}$
 from the previous section, whether the procedure outlined in Section~\ref{sec:estimates_d} delivers consistent estimates for $d$ or not, now depends on the specific model in question and on $g_\epsilon(d)$.

\begin{theorem}\label{theo:consistency_hat_d}
Consider the implicit estimators~\eqref{def:estimators_d_implicit}.
Assume that the conditions of Theorem~\ref{theo:rate_p1} required for the convergence of $\hat p_{n,\epsilon,1}$ and $\hat p_{n,2\epsilon,1}$ with rate $m_n^{-1/2}$ hold.
For that $\epsilon$, $d$, and $m_n$, assume that, as $n\to\infty$,
\begin{equation}\label{eq:bias}
p_{2\epsilon,1} = p_{\epsilon,1} \cdot g_\epsilon\Big\{d + o(m_n^{-1/2})\Big\}. \tag{B}
\end{equation}
Assume also that the derivative (with respect to $d$) of $g_\epsilon(d)$ exists, is continuous and non-zero at $d$.
If $p_{\epsilon,1}\cdot(1-p_{2\epsilon,1}) = o\big[n\{p_{\epsilon,2\epsilon,2}-p_{\epsilon,1} \cdot p_{2\epsilon,1}\}\big]$ and $m_n=o(n)$, then as $n\to\infty$,
\[
S_{n,\epsilon}^{-1/2} \cdot \left\{ \hat d_n - d \right\} \stackrel{d}{\longrightarrow}
N(0,\, 1), \qquad \hbox{where}\qquad
S_{n,\epsilon} = \left\{\frac{\partial\log g_\epsilon(d)}{\partial d}\right\}^2 \cdot \frac{\mathcal V_\epsilon}{m_n}
\]
where for $p_{\epsilon_1,\epsilon_2,2}=\mathbb{P}\{r(X,Z)\le\epsilon_1, r(Y,Z)\le\epsilon_2\}$,
\[
\mathcal V_\epsilon = %6\frac{m_n}n +
\frac{p_{\epsilon,1}^2 \cdot p_{2\epsilon,2} - 2 \cdot
p_{\epsilon,2\epsilon,2} \cdot p_{\epsilon,1} \cdot p_{2\epsilon,1}
+p_{2\epsilon,1}^2 \cdot p_{\epsilon,2}}{p_{\epsilon,1}^2 \cdot p_{2\epsilon,1}^2}
\]
\end{theorem} \hfill \BlackBox

\begin{remark}
In Theorem~\ref{theo:consistency_hat_d} we consider the case $m_n = o(n)$, which is the most relevant case in practice.
The general expression for $\mathcal{V}_\epsilon$ (meaning for any sequence $m_n\le n$) can be found in~\eqref{eq:variance_d_general}, in the Appendix.
\end{remark}

\begin{remark}
Condition~\eqref{eq:bias} controls the asymptotic bias of the estimator for $d$.
Note that this condition should be interpreted as a condition on $g_\epsilon$ and on the sequence $m_n$, and not a condition on $\epsilon$, since $\epsilon$ is a modelling parameter set by the user.
If condition~\eqref{eq:bias} does not hold, then the statement of the previous theorem is still valid if we centre $\hat d_n$ with $\mathbb{E}(\,\hat d_n\,)$ instead of $d$, but in that case the estimator might be asymptotically biased.
Alternatively, if~\eqref{eq:bias} holds with $r_n=o(m_n)$, instead of $m_n$, then we conclude that $r_n^{-1/2}\big(\hat d_n-d\big)= o_P(1)$.
(Note that the condition becomes more restrictive for faster rates.)
\end{remark}

For the explicit estimator in~\eqref{def:estimators_d_explicit} the $g_\epsilon(d)$-dependent scaling in the variance is $\log(2)^{-2}$.
In this case, the bias condition~\eqref{eq:bias} reduces to
\[
p_{2\epsilon,1} = p_{\epsilon,1} \cdot 2^{d+o\big(m_n^{-1/2}\big)}.
\]
This requires the connection probability $p_{\epsilon,1}$ to approximately have a doubling property:
if the distance at which vertices connect doubles, then the probability of connection goes up by a factor $2^d$, approximately.
(Note that also $\epsilon$ should in general depend on $n$, but we return to this point in Section~\ref{sec:simulations:scaling}.)
How much leverage we have in terms of the approximation depends mostly on the sequence $m_n$ and is, in a sense, the price to pay for the computational speed-up.
However, the choice of $m_n$ goes well beyond this.

\begin{remark}
Since our estimator for $d$ only requires access to $m_n$ rows of $\bm A_\epsilon$ and $\bm A_{2\epsilon}$, one can obtain several (correlated) estimates $\hat d_n^{(1)}$, $\hat d_n^{(2)}$, \dots, of $d$ based on disjoint sets of rows if $m_n = o(n)$.
From these one can estimate the variance of $\hat d_n$.

Say that we consider the following estimator for the variance
\[
\hat\sigma_n^2 =
\widehat{\mathbb{V} \hat d_n} =
\frac1{t_n-1}\sum_{i=1}^{t_n-1}\Big\{\hat d_n^{(i)} - \bar{d_n}\Big\}^2, \qquad\hbox{with}\qquad
\bar{d_n} = \frac1{t_n}\sum_{i=1}^{t_n} \hat d_n^{(i)},
\]
where $t_n \in \mathbb{N}$ is at most $n/m_n$.
It is straightforward to check that
\[
\mathbb{E}\hat\sigma_n^2 = \sigma^2 \Big[1-\rho\big\{\hat d_n^{(1)},\hat d_n^{(2)}\big\}\Big],
\]
where $\rho(X,Y)$ represents the correlation between $X$ and $Y$.
Making use of the fact that $\hat p_{n,\epsilon,1}^{(i)}$ and $\hat p_{n,2\epsilon,1}^{(i)}$ are independent for $i\in\mathbb{N}$, we have
\[
\mathbb{V}\big\{\hat d_n^{(1)},\hat d_n^{(2)}\big\} =
\frac1{\big\{\log(2)^2\big\}}\left[
\mathbb{V}\big\{\log \hat p_{n,\epsilon,1}^{(1)},  \log \hat p_{n,\epsilon,1}^{(2)}\big\} +
\mathbb{V}\big\{\log \hat p_{n,2\epsilon,1}^{(1)}, \log \hat p_{n,2\epsilon,1}^{(2)}\big\}\right].
\]
Using the approximation\footnote{
Here we use the fact that
$
\mathbb{V}(\log X, \log Y) =
\mathbb{V}\{\log(X/\mathbb{E}X), \log(Y/\mathbb{E}Y)\} \approx
\mathbb{V}(X/\mathbb{E}X-1, Y/\mathbb{E}Y-1) =
\mathbb{V}(X,Y)/(\mathbb{E}X\cdot\mathbb{E}Y)
$,
since $\log(1+x)\approx x$, for all appropriately small $x$.
} $\mathbb{V}(\log X, \log Y) \approx \mathbb{V}(X,Y)/(\mathbb{E}X\cdot\mathbb{E}Y)$, it is enough to look at $\mathbb{V}\big\{\hat p_{n,\epsilon,1}^{(1)}, \hat p_{n,\epsilon,1}^{(2)}\big\}$.
If we denote the range of rows associated with these two estimators as respectively $I_1$ and $I_2$, then writing the covariance as a four-fold sum, we get
\begin{align*}
\mathbb{V}\big\{\hat p_{n,\epsilon,1}^{(1)}, \hat p_{n,\epsilon,1}^{(2)}\big\} &=
\left\{ \frac2{m_n(n-1)}\right\}^2
\sum_{i_1\in I_1}\sum_{i_2\in I_2}\sum_{j_1=i_1+1}^n\sum_{j_2=i_2+1}^n
\mathbb{V}\big( A_{\epsilon,i_1,j_1}, A_{\epsilon,i_2,j_2} \big)\\ &\le
\frac4{n-1}\big(p_{\epsilon,2}-p_{\epsilon,1}^2\big),
\end{align*}
where we use the fact that since $I_1\cap I_2 = \emptyset$, then
$\mathbb{V}\big( A_{\epsilon,i_1,j_1}, A_{\epsilon,i_2,j_2} \big) = p_{\epsilon,2}-p_{\epsilon,1}^2$ if $j_1=j_2$, and
$\mathbb{V}\big( A_{\epsilon,i_1,j_1}, A_{\epsilon,i_2,j_2} \big) = 0$, if $j_1\neq j_2$.
Finally, under the conditions of Theorems~\ref{theo:rate_p1} and~\ref{theo:consistency_hat_d}, we can put everything together to bound
\[
\left|\mathbb{E}\hat\sigma_n^2 - \sigma^2\right| \lesssim
\frac{m_n}n \cdot \frac{S_{n,\epsilon,1}}{S_{n,\epsilon}};
\]
this upper bound converges to zero for appropriate $\epsilon$, if $m_n = o(n)$.
\end{remark}

Note that when the metric $r$ is induced by the Euclidean norm $\|\cdot\|_2$, and $m_n=n$, the estimator $\hat p_{n,\epsilon,1}$ (as a function of $\epsilon$) coincides with a realisation of the so-called correlation integral; cf.~\citep{camastra2002estimating}.
This is defined in the following way.
With $x_1,\dots, x_n$ denoting points on a manifold whose dimension we would like to measure, let
\begin{equation}\label{def:correlation_integral}
C_n(\epsilon) =
\frac2{n(n-1)}\sum_{i=1}^{n-1}\sum_{j=i+1}^n
\indic_{\{\|x_i-x_j\|_2\le \epsilon\}};
\end{equation}
the correlation integral $C(\epsilon)$ is the limit, when $n\to\infty$, of $C_n(\epsilon)$.
The underlying idea behind the intrinsic dimension being $d$ is that $C(\epsilon)$ should scale like $\epsilon^d$ so that the limit as $\epsilon\to 0$ of $\log\{C(\epsilon)\}/\log(\epsilon)$ is $d$;
this is then called the correlation dimension, which is a type of fractal dimension.
Following up on Remark~\ref{rem:p2vsp1}, basing our estimator on $p_{\epsilon,2}$ would lead to a variation on the correlation integral above.
We found no advantage in using the $p_{\epsilon,2}$ based estimator over the $p_{\epsilon,1}$ based one.

Some estimators for intrinsic dimension like that of~\citep{grassberger1983measuring} are based on the idea of regressing $\log\{C_n(\epsilon)\}$ on $\log(\epsilon)$, and estimating the (correlation) dimension from the slope of the fit.
However, based on the discussion from Section~\ref{sec:role_of_epsilon}, this slope actually corresponds to some average dimension over different scales, which is not what we are interested in; cf.~Fig.~3.3 of~\citealp{lee2007nonlinear}, for an example of the scale dependence of the correlation integral.
Further, this also makes extensive use of distances between the observations, while we rely only on the neighbourhood information provided by adjacency matrices.

As we argue in the previous section, one has to examine the data at (at least) two different scales to derive meaningful information about dimension, and this has been noted in the literature before.
\citep{kegl2002intrinsic} proposed a scale dependent notion of correlation dimension based on
\begin{equation}\label{eq:scale_dependent_correlation_dimension}
D_n(\epsilon_1,\epsilon_2) =
\frac{\log\{C_n(\epsilon_2)\} - \log\{C_n(\epsilon_1)\}}{\log(\epsilon_2)-\log(\epsilon_1)}
\end{equation}
This improves upon the idea of regressing the logarithms of the empirical version of the correlation integral (cf.~\citealp{grassberger1983measuring, pettis1979intrinsic}) on the logarithm of $\epsilon$ by allowing one to focus on a specific range of scales.
Indeed, this is closely reflected in our estimator $\hat d_n$, although we distinguish ourselves from other approaches by deriving the asymptotic distribution of our estimator within a flexible framework;
to the best of our knowledge the central limit theorem that we derive is new to the literature.

Other approaches to intrinsic dimension estimation follow similar ideas but make use of other notions of dimension like for example
Hausdorff dimension,
information dimension,
box counting dimension,
(generalised) expansion dimension,
packing dimension,
and also local versions of these concepts.
The success of such approaches depends mostly on how computationally tractable computing the estimate of the dimension is, and how adequate the particular notion of dimension at hand is for the model under consideration.

Another approach corresponds to the maximum likelihood estimator of~\citep{bickel2004some}.
This estimator is based on maximising the likelihood obtained by assuming that the observations come from a homogeneous Poisson process.
The estimator is then based on distances to the $k$-the nearest neighbour of each point, with $k$ interpreted as a ``bandwidth'' parameter of the estimator.
Another estimator based on $k$-nearest-neighbours is that of~\citep{kleindessner2015dimensionality}.
In both cases the connection between $k$ and the scale at which we estimate dimension has not yet been explored in detail, however.
The disadvantage of $k$-nearest-neighbour approaches seems to be that the way in which the distance of an observation to its $k$ nearest neighbour scales with dimensions may heavily depend on the underlying distribution of the data.
Although this does open the door to sharper estimates of the intrinsic dimension, this is done at the expense of needing more information about the distribution of the data.
Besides this, relating $k$ to the scale at which the intrinsic dimensions is being estimated also seems to be difficult.

The approaches mentioned above, as well as our approach, are examples of so called geometric methods.
A different class of methods are eigenvalue (or projection) methods. These stem from the work of~\citep{fukunaga1971algorithm}; see also~\citep{bruske1998intrinsic}.
These methods are typically based on principal component analysis (PCA) and estimate the dimension based on how many eigenvalues are above certain (small) threshold.
They seem to be less useful for estimating intrinsic dimension because of the difficulty of determining what constitutes an appropriate threshold; cf.~\citep{verveer1995evaluation}.

In the next section we present some numerical results to illustrate our approach.
These results guide us in our choice of the sequence $m_n$.

\section{Numerical Results}\label{sec:simulations}

In this section we present some numerical results.
We start by exemplifying in Section~\ref{sec:simulations:scaling} how the probabilities $p_{\epsilon,1}$ and $p_{\epsilon,2}$ determine the bias and variance of our estimator for different distributions for the observations $\bm X$.
Section~\ref{sec:simulations:choice_m_n} is about the consequences of the choice of the sequence $m_n$ in our estimator.
In Section~\ref{sec:simulations:combinations} the performance of our estimator is evaluated for different combinations of dimension $d$ and resolution $1/\epsilon$.
The main goal of these first three subsections is to understand what constitutes a good choice for the sequence $m_n$ that features in the definition~\eqref{def:p1_hat}.
Section~\ref{sec:simulation:better_g} concerns a non-trivial choice for the function $g_\epsilon$ (meaning a choice other than $2^d$), as well as other types of bias correction.
In Section~\ref{sec:simulation:effect_of_noise} we illustrate the effect of noise on our estimator.
Finally, in Section~\ref{sec:simulations:comparison}, we apply our estimator to a batch of data sets of both simulated, and real data.
To simplify the exposition, in all cases the metric $r$ is the Euclidean distance.

\subsection{Scaling of $p_{\epsilon,1}$ and $p_{\epsilon,2}$, and their Influence on the Bias and Variance of $\hat d_n$}\label{sec:simulations:scaling}

The probabilities $p_{\epsilon,1}$ and $p_{\epsilon,2}$ play an important role in our approach.
The quantity $\{p_{\epsilon,2}-p_{\epsilon,1}^2\}/p_{\epsilon,1}^2$ is the variance of the function $p_\epsilon(x)/p_{\epsilon,1}$, which is the relative connection probability at each site $x$.
We see, for example, that if $m_n=o(n)$, then the scaling $m_n\cdot S_{n,\epsilon}$ that features in the asymptotics for our estimator for $d$ is up to a constant factor the variance of $p_\epsilon(X)/p_{\epsilon,1}-p_{2\epsilon}(X)/p_{2\epsilon,1}$.
However, this quantity still depends on $\epsilon$ and $d$, so it is interesting to see how it behaves for different distributions for the design points.

\begin{figure}[!tb]
\centering
\includegraphics[width=1.00\textwidth]{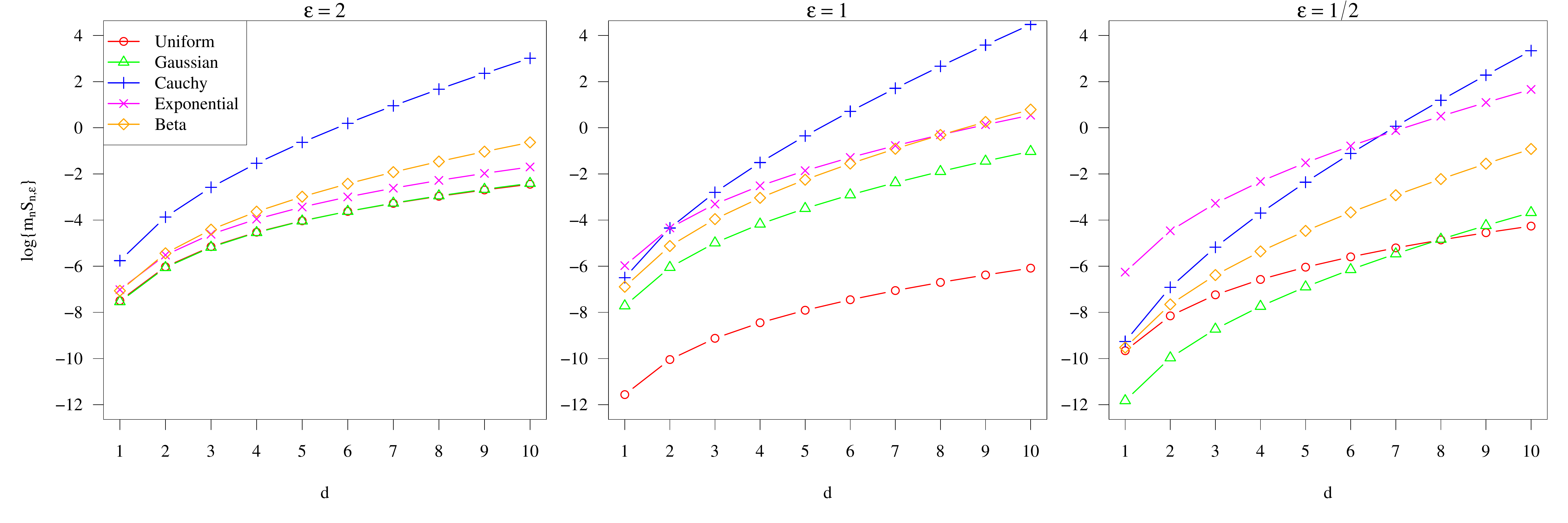}\caption{\footnotesize
Effect of the distribution of the design points, $d$, and $\epsilon$ on the logarithm of the asymptotic variance of our estimator for $d$.
The three plots correspond to $\epsilon\in\{2,1,1/2\}$, left to right.
In each plot, each line corresponds to a different distribution for the design points.
}\label{fig:scaling}
\end{figure}

In Figure~\ref{fig:scaling} we plot the logarithm of $m_n\cdot S_{n,\epsilon}$, as a function of the intrinsic dimension $d$ for several different choices for the distribution of the design points, for three choices of $\epsilon$.
If $\epsilon$ is fixed, then $m_nS_{n,\epsilon}$ is just the constant that features in the rate of convergence of the estimator of $\hat d_n$.
(The curves were computed by numerical integration, but are otherwise exact.)
The coordinates of the design points $X_i$ were sampled independently from the indicated distributions (uniform, Gaussian, exponential, and beta$\{2,5\}$, all scaled standard deviation $1$; Cauchy with scale parameter 1).
The main message is that for appropriately large $d$ (about $d\ge 5$) the lines increase roughly linearly, which would mean that the constants in the asymptotic statement in Theorem~\ref{theo:consistency_hat_d} increase exponentially with $d$.
In general, $d$ is fixed, but these plots give an indication that if the intrinsic dimension is large, then in order to attain a given level of precision, one should need a fairly large number of observations.
This should give a notion of when the asymptotics described in Section~\ref{sec:consistency_d_epsilon} \emph{kick in}.
The effect of $\epsilon$, on the other hand, does not seem too pronounced and affects mostly how the lines behave when the intrinsic dimension $d$ is relatively small.\par

\begin{figure}[!tb]
\centering
\includegraphics[width=1.00\textwidth]{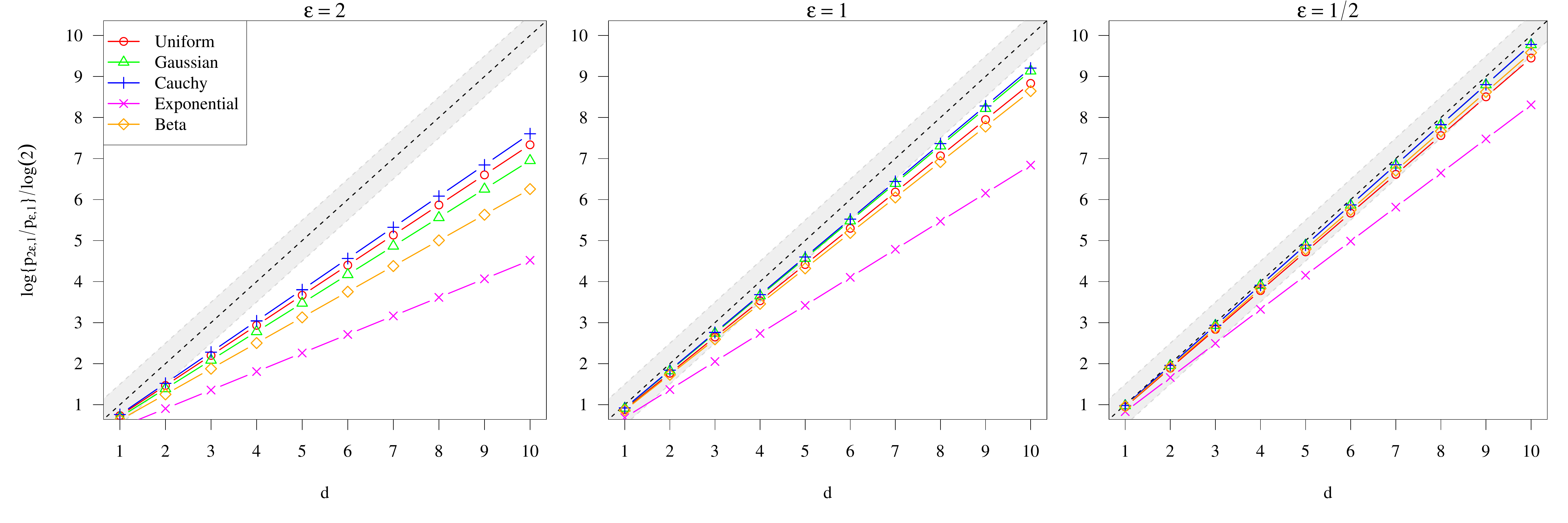}\caption{\footnotesize
Effect of the distribution of the design points, $d$, and $\epsilon$ on $\{\log p_{2\epsilon,1} - \log p_{\epsilon,1}\}/\log(2)$.
Left to right, the plots correspond to $\epsilon\in\{2,1,1/2\}$.
The shaded area corresponds to $d\pm1/2$.
}\label{fig:doubling}
\end{figure}

While the variance of the estimator is, up to the scaling $\{\partial\log g_\epsilon(d)/\partial d\}^2$, only model dependent, the bias depends greatly on the function $g_\epsilon$ used in the definition of the estimator.
In particular it depends on how well $g_\epsilon(d)$ approximates $p_{2\epsilon,1}/p_{\epsilon,1}$ as prescribed by the bias condition~\eqref{eq:bias}.
As discussed in Section~\ref{sec:estimates_d}, for appropriately small $\epsilon$, it should hold that $p_{2\epsilon,1}/p_{\epsilon,1}\approx 2^d$, making $g_\epsilon(d)= 2^d$ our canonical choice for $g_\epsilon$.
For this choice of $g$ and for the same distributions as before, in Figure~\ref{fig:doubling} we plot $d\mapsto \{\log p_{2\epsilon,1} - \log p_{\epsilon,1}\}/\log(2)$, for different $\epsilon$;
we compare it with the identity $d\mapsto d$.

In the plots above, the black dotted line along the diagonal is the identity, and the grey shaded area encompasses $d\pm 1/2$ (for reference).
As before, the remaining lines correspond to different distributions for the design points.
As expected from the discussion in Section~\ref{sec:estimates_d}, as $\epsilon$ gets smaller, these lines mimic the doubling property of the Lebesgue measure more closely so that $d\mapsto p_{2\epsilon,1}/p_{\epsilon,1}$ indeed gets closer to $d\mapsto 2^d$.
The plots also suggest that considering $g_\epsilon(d) = 2^d$ should lead to the dimension being systematically underestimated, and that one may want to consider a multiplicative correction based on $\epsilon$.
In a sense, this is the price to be paid for having a parameter-free estimator.

Part of the bias is a consequence of the fact that, in general, we do not have access to a good (model dependent) function $g$.
For example, if $g$ is constant over $\epsilon$, as discussed in Section~\ref{sec:estimates_d}, the bias should, in fact, be of order $\epsilon$.
One should therefore consider to take $\epsilon$ small to reduce the bias of the estimates.
However, this is somewhat at odds with the notion put forward in Section~\ref{sec:role_of_epsilon}, where we explain that $\epsilon$ determines the scale at which we examine the data and is therefore a (fixed) modelling parameter.

As it turns out, there is a justification for taking $\epsilon=\epsilon_n$ converging to zero.
The rationale is the following.
Typically, there will be noise in our data so that the support of $F$ is an enlarged version of the manifold whose dimension we would like to estimate.
If the design points are relatively concentrated, in the sense that $\mathbb{P}\big\{r(0,X)>x\big\}\lesssim\exp(-x^2)$, $x>0$, say, then, by the union bound,
\[
\mathbb{P}\Big\{\max_{i=1,\dots, n}r(0,X_i)>\sqrt{\delta\log n}\Big\}\le
\sum_{i=1}^n\mathbb{P}\Big\{r(0,X_i)>\sqrt{\delta\log n }\Big\} \lesssim
n^{1-\delta}, \qquad \delta>1.
\]
In other words, if $\epsilon_1$ is to express the distance at which we would like to analyse the data, then if we observe only $n$ points and take $\epsilon\equiv\epsilon_1$, we are overestimating the typical distances between points by roughly a factor $\sqrt{\log n}$.
In a sense, this growing spread can be thought of as arising from noise, so that we should therefore establish connections at a slightly smaller distance, say for instance $\epsilon_n=\epsilon_1/\sqrt{\log(1+n)}$, $\epsilon_1>0$.
This can also be seen as a finite sample correction for the estimator; cf.~\citep{grassberger1988finite}.
Another reason to consider $\epsilon$ of this kind would be to ensure that the adjacency matrices that we work with remain relatively sparse. This means that we avoid storage problems even when the sample size $n$ is large.
This can also be motivated from the point of view of discriminability; cf.~\citep{beyer1999nearest, weber1998quantitative, houle2013dimensionality}.
If the dimensionality of the data is high, then distance values are less discriminative, in the sense that they tend to concentrate more around the mean of their distribution.
Because of this, it makes sense to increase the strictness with which new connections are accepted as the sample size grows.\par

This has three  important consequences.
The first is that for the parameter-free estimator~\eqref{def:estimators_d_explicit}, with the finite sample correction described above should have squared bias $O(1/\log n)$.
This means that the sequence $m_n$ should be set to $O(\log n)$ to balance variance and squared bias.
The proverbial \emph{less is more} comes to mind:
picking $m_n$ large and averaging over many vertices leads to deceptive results, since the variance of the estimate is reduced, while the bias remains unchanged.
This is an inherent feature of estimators obtained via inversion, but it is something that is invariably missed in the literature --
estimates are strongly concentrated around biased estimates.
This is undesirable from the point of view of uncertainty quantification; see also the next section.
By doing this our estimator attains the minimax rate for this problem which is known to be logarithmic;
cf.~\citep{koltchinskii2000empirical}.

The second consequence follows from the first:
setting $m_n = O(\log n)$ leads to an algorithm with complexity $O(n\log n)$.
This is a great advantage over competing algorithms whose execution time typically scales like $O(n^2)$, sometimes like $O(D\, n^2)$; cf.~Table 1 in~\citep{eriksson2012estimating}.
Finally, since $m_n$ is rather small compared with $n$, this means that we are estimating $d$ based on the degrees of only a few vertices.
By repeating the estimation for disjoint sets of vertices we can estimate the standard deviation of the estimator without a need for resampling.
The conclusion is that picking small $m_n$ is better, both from a theoretical and practical perspective.

In the next subsection we perform some numerical experiments to investigate more closely the consequences of different choices for $m_n$.

\subsection{Different Choices of $m_n$}\label{sec:simulations:choice_m_n}

\begin{figure}[!tb]
\centering
\includegraphics[trim={5.5cm 9cm 5.5cm 8cm},clip, width=0.50\textwidth]{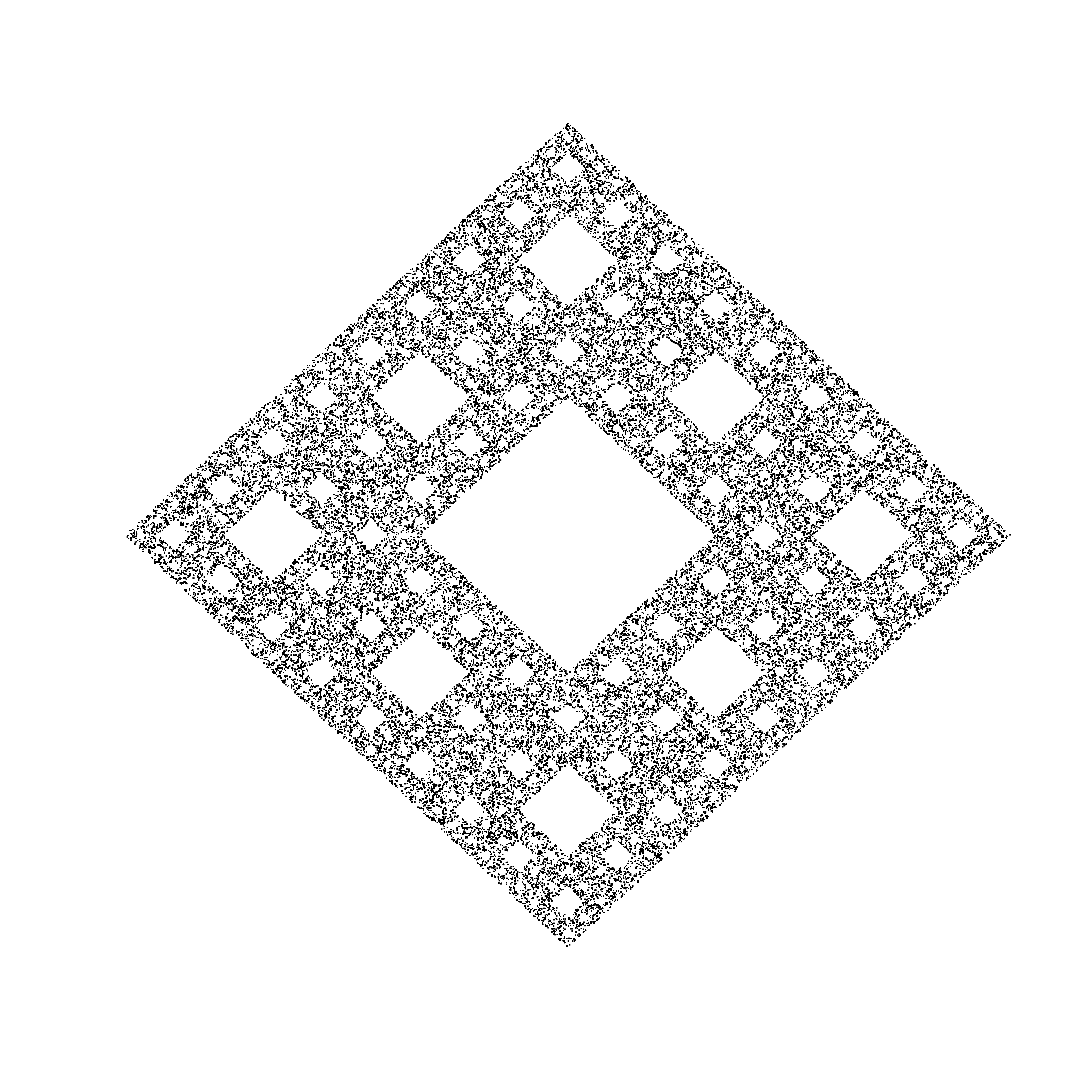}\caption{\footnotesize
Example of $5\cdot 10^4$ points sampled uniformly at random on a Sierpinski carpet.
Points like these are used as design points in the simulations in this section.
}\label{fig:sierpinski}
\end{figure}

In this section we look more closely at the choice of $m_n$ by exemplifying the effect that the choice of this sequences has on:
a) the bias,
b) the variance, and
c) the execution time.
To have a nontrivial dimension we consider design points sampled uniformly at random on a Sierpinski carpet.
This can be done in the following way.
Consider
\[
P_0 = \begin{bmatrix} 0\\ 0\end{bmatrix}, \qquad\hbox{and}\qquad
\bm C = \begin{bmatrix*}[r]
0   &1/2  &1  &1/2  &0  &-1/2  &-1  &-1/2\\
1   &1/2  &0  &-1/2 &-1 &-1/2  &0   &1/2
\end{bmatrix*}.
\]
Let $e_i=[0\, \cdots\, 0\, 1\, 0\, \cdots\, 0]^T$, $i=1,\dots,8$, be unit vectors that have a $1$ in the $i$-th position, and let $i_j\sim U\{1,\dots,8\}$, $j=1,2,\dots$, be a sequences of independent, discrete uniform random variables taking values on $\{1,\dots,8\}$.
A point can be drawn uniformly at random on a Sierpinski carpet as
\[
P = P_0 + \bm C \sum_{j=1}^\infty 3^{-j} \cdot e_{i_j}.
\]
Figure~\ref{fig:sierpinski} depicts $5\cdot10^4$ points drawn according to this procedure.
In practice we truncate the sum at $100$ terms.
(Note that this is accurate enough to get the neighbourhood matrices $\bm A_\epsilon$ exactly.)
The correlation dimension of the Sierpinski carpet is, to the best of our knowledge, unknown, but its Hausdorff dimension is $\log8/\log3\approx 1.89$, which should provide a good indication to what the intrinsic dimension should be.

In this section we ran our algorithm for all combinations of $n=10\cdot2^\gamma$, $\gamma\in\{0,1,\dots,8\}$, and $m_n$ either $\log(n)$ or $n^\gamma$, $\gamma\in\{1/4,1/2,3/4,1\}$.
The parameter $\epsilon$ was set to $\epsilon_n = {\rm sd}\{{\rm vect}(\bm X)\}/\sqrt{\log(n+1)}$, where ${\rm vect}(\bm X)$ represents the entries for $\bm X$ written as a vector;
the standard deviation of the coordinates of the design points, although in practice unknown, was used here to ensure that $\epsilon$ is on the right scale.
These simulations were repeated $10^5$ times and the results are averaged.

{\renewcommand{\arraystretch}{1.3}
\begin{table}[!tb]
\centering
\begin{tabular}{crccccc}
& \multicolumn{1}{c}{} 	& \multicolumn{5}{c}{$m_n$}                            \\ \cline{3-7}
& \multicolumn{1}{c}{}
	      & $\log(n)$   		  &  $n^{1/4}$  		  & $n^{1/2}$   	   	& $n^{3/4}$ 		  	& $n$ \\ \hline\hline
\multirow{8}{*}{$n$}
& 10	  & 1.1780 (0.78)	& 1.1243 (0.55)	& 1.0945 (0.39)	& 1.0877 (0.32)	& 1.0861 (0.26)\\
& 20	  & 1.2678 (0.54)	& 1.2528 (0.43)	& 1.2483 (0.33)	& 1.2409 (0.24)	& 1.2367 (0.18)\\
& 30	  & 1.3784 (0.49)	& 1.3718 (0.40)	& 1.3673 (0.26)	& 1.3631 (0.17)	& 1.3602 (0.12)\\
& 40	  & 1.4852 (0.44)	& 1.4781 (0.35)	& 1.4668 (0.20)	& 1.4660 (0.12)	& 1.4636 (0.08)\\
& 160	  & 1.5491 (0.32)	& 1.5441 (0.27)	& 1.5434 (0.15)	& 1.5398 (0.08)	& 1.5395 (0.05)\\
& 320	  & 1.6037 (0.28)	& 1.5963 (0.22)	& 1.5925 (0.12)	& 1.5930 (0.06)	& 1.5923 (0.03)\\
& 640	  & 1.6395 (0.27)	& 1.6376 (0.19)	& 1.6332 (0.09)	& 1.6320 (0.04)	& 1.6320 (0.02)\\
& 1280	& 1.6595 (0.22)	& 1.6622 (0.18)	& 1.6596 (0.07)	& 1.6595 (0.03)	& 1.6597 (0.01)\\
& 2560  & 1.6790 (0.22)	& 1.6770 (0.15)	& 1.6778 (0.06)	& 1.6768 (0.02)	& 1.6772 (0.01)\\ \hline
\end{tabular}
\caption{\footnotesize
Results for the estimation of the intrinsic dimension $d$ for random Sierpinski carpet design points, for different combinations of $n$ and $m_n$.
For each combination, $\hat d_n$ was estimated $10^5$ times;
we display the mean estimate, and in parenthesis the standard deviation among the estimates.
}\label{table:n_vs_m_n}
\end{table}
}

Table~\ref{table:n_vs_m_n} summarises the average and standard deviation of the estimates that were obtained for each combination of $n$ and $m_n$.
Irrespectively of the sequence $m_n$, it is clear that as $n$ grows the estimates stabilise.
This is in tune with our consistency result, also in that the reduction of the bias is rather slow.
It is also clear from the results that the sequence $m_n$ does not seem to have much influence on the quality of the estimate, particularly as $n$ grows.
This is also in tune with our results:
larger $m_n$ does increase the precision of the estimates of the probabilities $p_{\epsilon,1}$;
what mainly determines the precision of the estimate of $d$ is the bias introduced by the function $g_\epsilon$, though.
The effect of $m_n$ on the standard deviation is also as expected:
increasing either $n$ or $m_n$ generally leads to a decreased of the variability of the estimate.
From this it might seem reasonable to set $m_n$ to a large value (after all, it does reduce the variance of the estimator without reducing precision).
There are however two good reasons to keep the growth of $m_n$ slow.

\begin{figure}[!tb]
\centering
\includegraphics[trim={0cm 0cm 0cm 0cm},clip, width=0.50\textwidth]{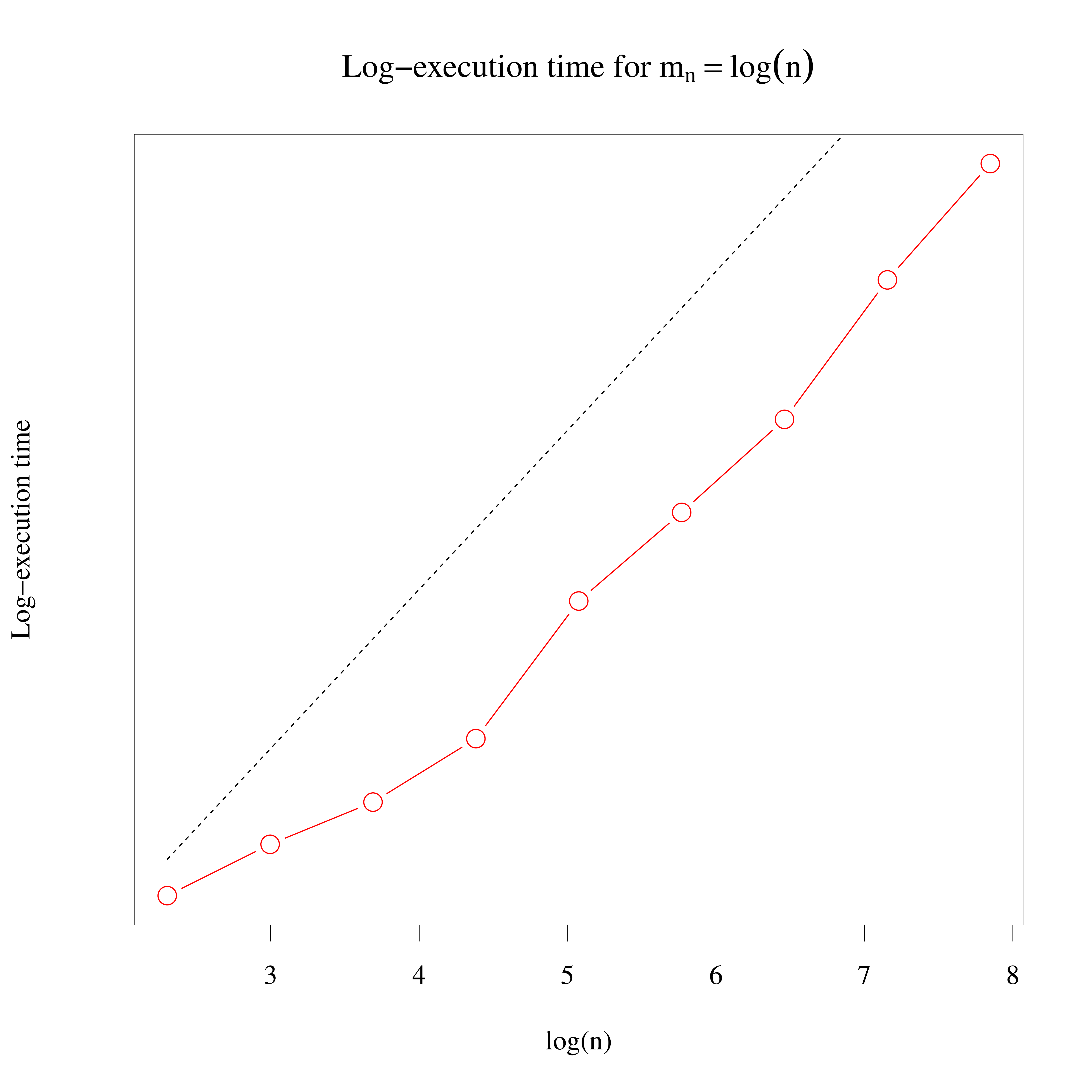}\caption{\footnotesize
The evolution of the average execution time of our algorithm as a function of $n$ when $m_n=\log(n)$.
The plot is on a $\log$-$\log$ scale.
For reference, the dashed line represents $\log(n)$ plotted against $\log(n) -10$.
The execution time grows roughly linearly with $n$.
}\label{fig:execution_time_log}
\end{figure}

The first reason is execution time.
Figure~\ref{fig:execution_time_log} shows the evolution of the average execution time of the algorithm as a function of $n$, when $m_n=\log(n)$.
For comparison, Table~\ref{table:execution_time} shows the average execution time as a multiplier of the execution time for $m_n=\log(n)$, for the same value of $n$.
In words: the numbers on the table indicate how much slower it is to set $m_n$ to each choice, compared to just setting it to $\log(n)$.

{\renewcommand{\arraystretch}{1.3}
\begin{table}[!tb]
\centering
\begin{tabular}{crcccc}
& \multicolumn{1}{c}{} 	& \multicolumn{4}{c}{$m_n$} \\ \cline{3-6}
& \multicolumn{1}{c}{}
	      &  $n^{1/4}$  		  & $n^{1/2}$   	   	& $n^{3/4}$ 		  	& $n$ \\ \hline\hline
\multirow{8}{*}{$n$}
& 10	  & 1.13 	& 1.65 	& 2.14	& 3.20\\
& 20	  & 1.29	& 1.63 	& 2.94	& 5.98\\
& 40	  & 1.34 	& 2.63 	& 5.59	& 12.72\\
& 80	  & 1.31 	& 3.51 	& 9.75	& 28.43\\
& 160  	& 1.20 	& 3.57 	& 12.10	& 42.46\\
& 320  	& 1.52 	& 5.24 	& 21.72	& 90.87\\
& 640  	& 1.91 	& 8.05 	& 39.19	& 194.22\\
& 1280  & 1.53	& 9.03 	& 53.86	& 325.12\\
& 2560	& 1.99 	& 12.68 & 89.80	& 638.54\\ \hline
\end{tabular}
\caption{\footnotesize
Average execution times for the algorithm.
For each combination of $n$ and sequence $m_n$, the respective entry in the table specifies how many times larger the execution time of the algorithm is compared to using $m_n=\log(n)$.
For example, if we set $m_n=n$, then, when $n$ is $2560$, we have to wait more than $638$ times longer for the algorithm to terminate than if we had used $m_n=\log(n)$.
}\label{table:execution_time}
\end{table}
}

The conclusion from Figure~\ref{fig:execution_time_log} is that the execution time for $m_n=\log(n)$ grows roughly linearly with $n$.
On the other hand, from Table~\ref{table:execution_time}, for other choices of $m_n$ the execution time quickly becomes prohibitive, particularly for faster-growing sequences $m_n$.

{\renewcommand{\arraystretch}{1.3}
\begin{table}[!tb]
\centering
\begin{tabular}{crccccc}
& \multicolumn{1}{c}{} 	& \multicolumn{5}{c}{$m_n$}                            \\ \cline{3-7}
& \multicolumn{1}{c}{}
	      & $\log(n)$   		  &  $n^{1/4}$  		  & $n^{1/2}$   	   	& $n^{3/4}$ 		  	& $n$ \\ \hline\hline
\multirow{8}{*}{$n$}
& 10	  & 84.75 	& 71.68 	& 47.59 	& 30.78 	& 15.16\\
& 20	  & 80.44 	& 69.40 	& 52.01 	& 24.14 	&  5.62\\
& 40	  & 83.08 	& 76.07 	& 48.46 	& 14.38 	&  0.60\\
& 80	  & 86.30 	& 79.55 	& 43.72 	&  6.03 	&  0.00\\
& 160	  & 82.07 	& 76.13 	& 36.11 	&  1.45 	&  0.00\\
& 320	  & 83.70 	& 73.95 	& 27.07 	&  0.12 	&  0.00\\
& 640	  & 85.73 	& 73.30 	& 17.81 	&  0.00 	&  0.00\\
& 1280	& 83.21 	& 75.57 	& 10.24 	&  0.00 	&  0.00\\
& 2560	& 84.95 	& 71.52 	&  4.52 	&  0.00 	&  0.00\\ \hline
\end{tabular}
\caption{\footnotesize
Percentage of the $10^5$ runs where the true value of $d$ is within $2$ standard deviations of $\hat d_n$.
}\label{table:confidence_interval}
\end{table}
}

There is a second reason to set $m_n=O\{\log(n)\}$.
Table~\ref{table:confidence_interval} shows the percentage of runs in which $d$ falls within $2$ standard deviations from $\hat d_n$.
The effect of $m_n$ is clear.
The confidence interval contains the true value of $d$, only when $m_n$ grows appropriately slowly.
Since the estimates are biased, if the sequence $m_n$ grows too quickly, then the variance of the estimates is too small.
The estimate $\hat d_n$ becomes overly concentrated around its biased mean.
In effect, because the bias and the variance are out of balance, the standard deviation fails to properly quantify the uncertainty in the estimate.
Also remember that if $m_n$ is small, then we can produce several estimates of $d$, from which we can estimate the standard deviation without a need for resampling.

To conclude, the sequence $m_n$ affects the variance of the estimator $\hat d_n$, but it does not affect the bias of the estimate, which comes mostly from the function $g_\epsilon$.
Faster growing $m_n$ therefore leads to estimates that are overly concentrated around their (biased) mean, so that their variability gives misleading information about the uncertainty in the estimate.
Such choices of $m_n$ also lead to a large computations cost.
Therefore, setting $m_n = O\{\log(n)\}$ is arguably the correct choice to make.

\subsection{Different Combinations of $d$ and $n$}\label{sec:simulations:combinations}

In this section we show how the estimator $\hat d_n$ behaves for different combinations of $d$ and $n$.
We set the distribution of the design points $X\sim N_d(0, \bm I)$, for $d \in \{1,2,3,4,5,10\}$, and chose $n\in\{10^3, 10^4, 10^5, 10^6, 10^7\}$;
irrespectively of the dimension we always set $\epsilon=\epsilon_n=4/(\log n)^{1/2}$.
Based on the discussion from the previous section, the parameter $m_n$ was set to $\max(1,\, \log n)$.

Table~\ref{table:combinations} below contains the results of estimating the intrinsic dimension $d$ $10$ times:
for each combination of $n$ and $d$ we sampled an adjacency matrix, estimated $d$ 10 times from 10 disjoint subsets of $m_n$ vertices (chosen at random, without replacement); the average and (in brackets) the standard deviation of the $10$ estimates make up the entries of the table.
Note that since we only average over a relatively small number of vertices in the graph, producing Table~\ref{table:combinations} does not actually require any resampling;
the same data set can be used (for each combination of $n$ and $d$).
Note also that for the choice of $m_n$ above, the execution time of the algorithm is $O(n\log n)$, meaning that it is considerable faster than any alternative approach.

{\renewcommand{\arraystretch}{1.3}
\begin{table}[!tb]
\centering
\begin{tabular}{crccccc}
& \multicolumn{1}{c}{} 	& \multicolumn{5}{c}{$n$}                            \\ \cline{3-7}
& \multicolumn{1}{c}{}
	& $10^3$   		& $10^4$  		& $10^5$   		& $10^6$ 			& $10^7$ \\ \hline\hline
\multirow{6}{*}{$d$}
& 1 	& 0.49 (0.15)	& 0.58 (0.11)	& 0.54 (0.12)	& 0.63 (0.07)	& 0.71 (0.15)\\
& 2	  & 0.91 (0.18)	& 1.18 (0.19)	& 1.31 (0.26)	& 1.41 (0.25)	& 1.53 (0.17)\\
& 3	  & 1.76 (0.24)	& 1.99 (0.29)	& 2.23 (0.36)	& 2.28 (0.34)	& 2.51 (0.20)\\
& 4	  & 2.55 (0.50)	& 2.94 (0.30)	& 3.23 (0.42)	& 3.05 (0.44)	& 3.42 (0.32)\\
& 5  	& 3.24 (0.37)	& 3.61 (0.38)	& 3.96 (0.43)	& 4.31 (0.18)	& 4.39 (0.44)\\
& 10	& 5.95 (0.61)	& 7.46 (0.53)	& 8.59 (0.70)	& 8.81 (0.70)	& 8.97 (0.75)\\ \hline
\end{tabular}
\caption{\footnotesize
Results for the estimation of $d$ for Gaussian design points for different combinations of $d$ and $n$.
For each combination $\hat d_n$ was estimated 10 times;
we display the mean estimate, and in parentheses the standard deviation among the estimates.
}\label{table:combinations}
\end{table}
}

A few things are clear from the results in Table~\ref{table:combinations}.
As hinted in Section~\ref{sec:simulations:scaling}, the estimator tends to underestimate the true intrinsic dimension, especially if the dimension is large.
However, as far as recuperating the integer dimension, the estimator performs well, especially considering that the data is entirely comprised of random fluctuations.
The standard deviation also does a good job at quantifying the precision of the estimate.

We emphasise that the estimates are parameter free --
one can improve the results with extra knowledge about the distribution of the data.
We do this in the following subsection.

\subsection{Non-canonical Choice of $g_\epsilon(d)$, and Bias Corrections}\label{sec:simulation:better_g}

In this section we propose some modifications of our estimator aimed at removing (or at least mitigating) its bias.
Since we know that the estimator $\hat d_n$ systematically underestimates $d$, a simple way of obtaining a ``bias corrected'' estimator is by scaling $\hat d_n$ up.
We consider three different ways of doing this.

Following the discussion at the end of Section~\ref{sec:estimates_d}, where we justified that the bias should be of order $O(d\cdot\epsilon)$, we consider
\begin{equation}\label{def:estimators_d_bias_correction}
\tilde d_n = \hat d_n \cdot \left\{ 1 + \frac2{\log(2)}\cdot \epsilon\right\}\cdot
\end{equation}
The motivation for considering this form is indeed the heuristic upper bound on the bias from Section~\ref{sec:estimates_d}.
We replaced the factor $3/\log(2)$ with a $2/\log(2)$ to be conservative, since we do not expect the upper bound on the bias to be tight.
We also use the fact that the ratio of expectations in the bound can reasonably be expected to be of order $d$ since, on average, the gradient of the density should only be non-trivial along $d$ independent directions.
A multiplicative correction of this form seems to be appropriate not just for the Gaussian case, where the expectations that feature in the bound on the bias can be computed.
Indeed, this form of the estimator performs quite well without any adjustments on subsequent experiments; see Table~\ref{table:simulations}.
With more knowledge about $F$ one can certainly make more informed choices about the specific constants in~\eqref{def:estimators_d_bias_correction}.
Table~\ref{table:combinations_corrected} shows the values that the bias corrected estimates~\eqref{def:estimators_d_bias_correction} take.

{\renewcommand{\arraystretch}{1.3}
\begin{table}[!tb]
\centering
\begin{tabular}{crccccc}
& \multicolumn{1}{c}{} 	& \multicolumn{5}{c}{$n$}                            \\ \cline{3-7}
& \multicolumn{1}{c}{}
	& $10^3$   	& $10^4$  	& $10^5$   	& $10^6$ 	& $10^7$ \\ \hline\hline
\multirow{6}{*}{$d$}
& 1	& 0.55 (0.17)	& 0.64 (0.12)	& 0.59 (0.13)	& 0.69 (0.08)	&  0.78 (0.16)\\
& 2	& 1.03 (0.20)	& 1.32 (0.21)	& 1.44 (0.29)	& 1.54 (0.28)	&  1.66 (0.18)\\
& 3	& 2.00 (0.27)	& 2.22 (0.32)	& 2.46 (0.40)	& 2.49 (0.37)	&  2.72 (0.22)\\
& 4	& 2.89 (0.57)	& 3.28 (0.34)	& 3.55 (0.46)	& 3.34 (0.48)	&  3.71 (0.35)\\
& 5	& 3.67 (0.42)	& 4.02 (0.43)	& 4.37 (0.48)	& 4.71 (0.19)	&  4.77 (0.48)\\
& 10	& 6.74 (0.69)	& 8.31 (0.59)	& 9.47 (0.77)	& 9.63 (0.77)	&  9.74 (0.81)\\ \hline
\end{tabular}
\caption{\footnotesize
This table contains the results of repeating the experiment from the previous section, but we compensate for the bias by considering a multiplicative correction.
}\label{table:combinations_corrected}
\end{table}
}
Comparing these results with those of Table~\ref{table:combinations}, we see that indeed this correction seems to substantially improve the estimates.

One can also consider other choices for the function $g_\epsilon$ to reduce the bias.
The idea is to use knowledge of the distribution of the design points to select a better suited candidate for this function.
This also leads to more precise estimates of the intrinsic dimension.
If $X\sim N_d(0, \bm I)$, independent of $Y\sim N_d(0, \bm I)$ then $Z=X-Y\sim N_d(0, 2\bm I)$, so that,
if we abbreviate $\mathcal Z_\epsilon = \{z\in\mathbb{R}^d: \|z\|\le\epsilon\}$, then
\[
p_{\epsilon,1} =
\mathbb{P}(\|X-Y\| \le \epsilon) =
\mathbb{P}(\|Z\| \le \epsilon) =
\frac1{(4\pi)^{d/2}}\int_{\mathcal Z_\epsilon} e^{-\frac14\|z\|^2}\, dz =
\frac{v_\epsilon}{(4\pi)^{d/2}}\int_0^1 e^{-\frac14u^2\epsilon^2}\, du,
\]
where $v_\epsilon$ represents the volume of a $d$-dimensional Euclidean ball of radius $\epsilon$.
The integral above can be expressed in terms of the Gauss error function $\rm erf$, so that
\[
p_{\epsilon,1} = (4\pi)^{-d/2}\cdot v_\epsilon\cdot \frac{\sqrt{\pi}}\epsilon\cdot {\rm erf}(\epsilon/2),
\qquad\hbox{whence}\qquad
\frac{p_{2\epsilon,1}}{p_{\epsilon,1}} =
2^d\cdot \frac{\rm erf(\epsilon)/2}{\rm erf (\epsilon/2)}\cdot
\]
What is arguably the ideal choice for the function $g_\epsilon$ is then
\[
g_\epsilon(d) =
2^d\cdot \frac{\rm erf(\epsilon)/2}{\rm erf (\epsilon/2)},
\qquad\hbox{leading to}\qquad
\bar d_n = \hat d_n +
\frac{\log\{{\rm erf(\epsilon/2)}\} - \log\{{\rm erf(\epsilon)/2}\}}{\log 2},
\]
where $\hat d_n$ is the canonical estimator from~\eqref{def:estimators_d_explicit}.
Note that using the exact function $g_\epsilon(d)$ does not entirely remove the bias of the estimate since $g_\epsilon(d)$ is not linear in $d$, and since $\hat p_{n,2\epsilon,1}/\hat p_{n,\epsilon,1}$ is not an unbiased estimator for $p_{2\epsilon,1}/p_{\epsilon,1}$.
Note also that the correction factor depends only on $\epsilon$, but not on $d$.
To understand the effect of this new estimator based on the more precise choice of $g_\epsilon(d)$, we repeat the numerical experiment of the previous section now for the estimator $\bar d_n$ from the previous display;
Table~\ref{table:combinations_better_g} summarises these results.
Comparing Tables~\ref{table:combinations} and~\ref{table:combinations_better_g}, it is clear that, as one would expect, a more informed choice for mapping $g_\epsilon$ considerably improves the estimates.
The improvement provided by this additive correction typically performs somewhat worse than the multiplicative correction, but the difference seems to be small.
This is most likely due to the fact that the multiplicative correction is better at compensating for the bias induced by the bias of our moment estimator of $p_{2\epsilon,1}/p_{\epsilon,1}$.

{\renewcommand{\arraystretch}{1.3}
\begin{table}[!tb]
\centering
\begin{tabular}{crccccc}
& \multicolumn{1}{c}{} 	& \multicolumn{5}{c}{$n$}\\ \cline{3-7}& \multicolumn{1}{c}{}
	  & $10^3$   		& $10^4$  		& $10^5$   		& $10^6$ 			& $10^7$ \\ \hline\hline
\multirow{6}{*}{$d$}
& 1	& 1.04  (0.25)	& 1.04  (0.11)	& 1.06  (0.14)	& 1.00  (0.11)	& 0.99  (0.15)\\
& 2	& 1.63  (0.22)	& 1.76  (0.33)	& 1.69  (0.24)	& 1.80  (0.27)	& 1.84  (0.17)\\
& 3	& 2.42  (0.37)	& 2.45  (0.48)	& 2.54  (0.23)	& 2.48  (0.33)	& 2.73  (0.33)\\
& 4	& 2.99  (0.41)	& 3.23  (0.35)	& 3.71  (0.38)	& 3.64  (0.32)	& 3.50  (0.28)\\
& 5	& 3.98  (0.54)	& 4.37  (0.47)	& 4.49  (0.44)	& 4.36  (0.34)	& 4.30  (0.33)\\
& 10& 6.56  (0.67)	& 7.87  (0.60)	& 8.47  (0.79)	& 9.63  (0.82)	& 9.39  (0.71)\\ \hline
\end{tabular}
\caption{\footnotesize
This table contains the results of repeating the experiment from the previous section, but instead using the true underlying function $g_\epsilon(d)$ that maps $d$ to the ratios $p_{2\epsilon,1}/p_{\epsilon,1}$.
}\label{table:combinations_better_g}
\end{table}
}

As a third and final alternative, one can also shift the estimates up by a factor depending on the standard deviations of the estimates.
This is motivated by the fact that our choice of $m_n = O\{\log(n)\}$ balances squared bias and variance.
(Again, the standard deviation is estimated using the same data as $\hat d_n$ without any need from resampling.)
Table~\ref{table:combinations_variance_correction} presents the results of adding (since we otherwise under-estimate) two standard deviations to the corresponding estimate $\hat d_n$.

{\renewcommand{\arraystretch}{1.3}
\begin{table}[!tb]
\centering
\begin{tabular}{crccccc}
& \multicolumn{1}{c}{} 	& \multicolumn{5}{c}{$n$}\\ \cline{3-7}& \multicolumn{1}{c}{}
	  & $10^3$   		& $10^4$  		& $10^5$   		& $10^6$ 			& $10^7$ \\ \hline\hline
\multirow{6}{*}{$d$}
& 1 	& 0.79	& 0.80	& 0.78	& 0.77	& 1.01\\
& 2	  & 1.27	& 1.56	& 1.83	& 1.91	& 1.87\\
& 3	  & 2.24	& 2.57	& 2.95	& 2.96	& 2.91\\
& 4	  & 3.55	& 3.54	& 4.07	& 3.93	& 4.06\\
& 5  	& 3.98	& 4.37	& 4.82	& 4.67	& 5.27\\
& 10	& 7.17  & 8.52	& 9.99	& 10.21	& 10.47\\ \hline
\end{tabular}
\caption{\footnotesize
This table contains the results of repeating the experiment from the previous section, but we compensate for the bias by adding two standard deviations to the estimate.
}\label{table:combinations_variance_correction}
\end{table}
}
This correction performs well, particularly considering that it requires no extra information about the distribution of the data.
Overall this seems to present the best correction for exactly this reason.

Although asymptotically any of the three corrections performs equally well, for finite samples, and with extra knowledge of the distribution of the design points, the multiplicative correction provides the best results.
Without extra knowledge though, the variance based correction still improves the canonical estimate $\hat d_n$ considerably.
However, this does not change the fact that one will always need a large sample size $n$ when the intrinsic dimension $d$ is large.

\subsection{Effect of the Noise}\label{sec:simulation:effect_of_noise}

\begin{figure}[!tb]
\centering
\includegraphics[				width=0.35\textwidth]{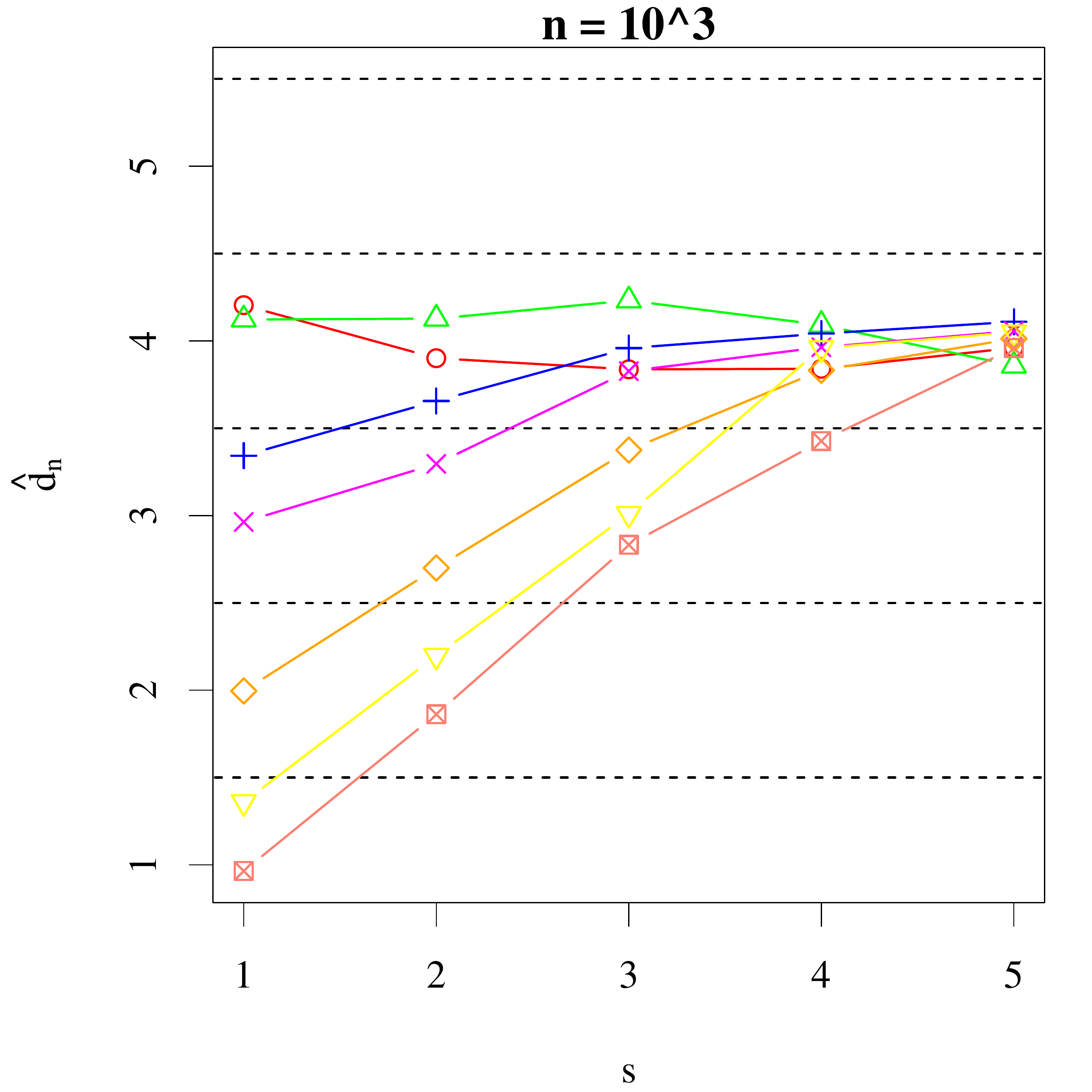}
\includegraphics[clip, trim={3cm 0 0 0},	width=0.31\textwidth]{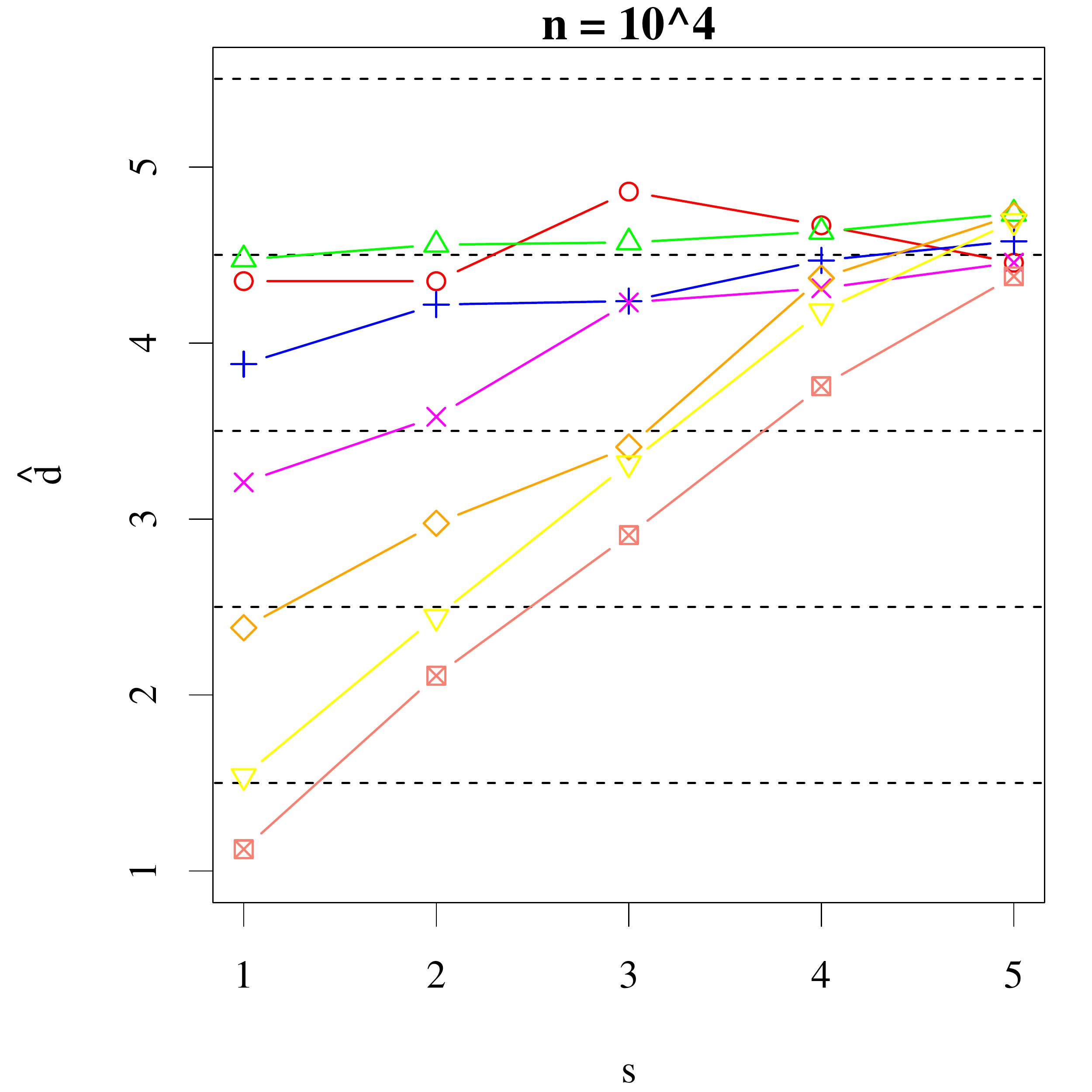}
\includegraphics[clip, trim={3cm 0 0 0},	width=0.31\textwidth]{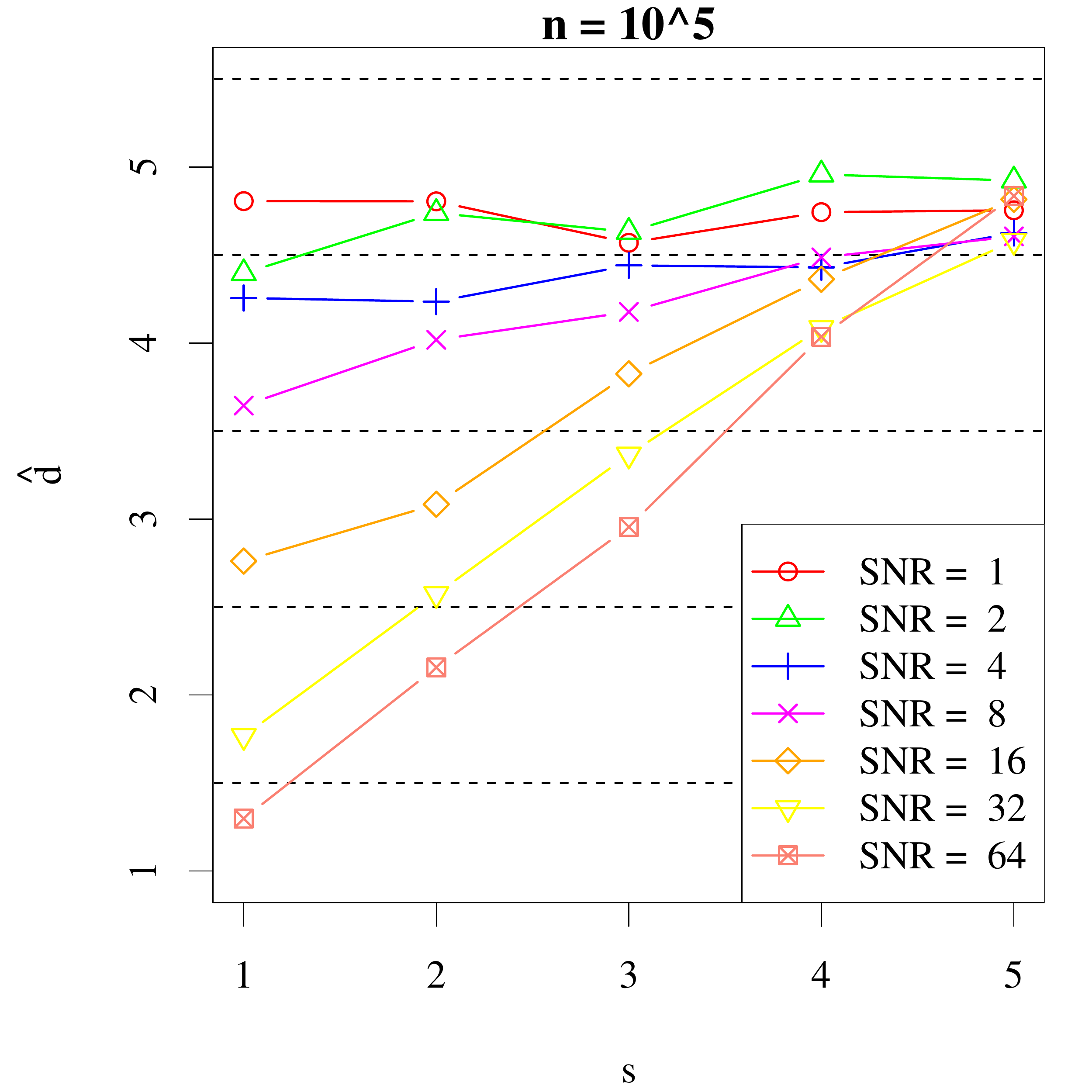}
\caption{\footnotesize
The estimator's sensitivity to noise.
Each plot corresponds to a different sample size, from left to right: $n\in\{10^3,10^4,10^5\}$.
The lines correspond to different estimates of the intrinsic dimension for different values of $s$,
and the different colours correspond to different SNR.
}\label{fig:eigen}
\end{figure}

An important aspect to be taken into consideration is the robustness of the estimation procedure to the presence of noise.
Irrespectively of the nature of the observations, if they are corrupted with enough noise, the estimation procedure will only detect the noise.
In this respect we would like to see how sensitive the estimator is to the presence of noise.
To see if this, we sampled design points $\bm X^{(s)}$ in $\mathbb{R}^5$ according to
\[
X_i^{(s)} \stackrel{i.i.d.}{\sim} N_5\Big\{ 0,\, \diag\big(
\overbrace{\sigma_{\rm{signal}}^2,.\,}^{s}.\,
\overbrace{.,\sigma_{\rm{noise}}^2}^{5-s}\big)\Big\}, \quad i=1,\dots, n,\; s=1, \dots,5.
\]
We set $\epsilon=\epsilon_n=\sigma_{\rm{signal}}(2\log n)^{-1/2}$, and ran our estimation procedure on the adjacency matrix obtained from $\bm X^{(s)}$ for $s=1,\dots,5$.
Each plot in Figure~\ref{fig:eigen} corresponds to a different sample size $n\in\{10^3,10^4,10^5\}$.
The coloured lines in each plot correspond to the 5 estimates $\hat d_n$, averaged over 10 disjoint sets of vertices.
Different colours correspond to different signal to noise ratio (SNR);
namely, we fixed $\sigma_{\rm{signal}}=1$ and chose $\sigma_{\rm{noise}}$ such that $\hbox{SNR} = \sigma_{\rm{signal}}^2/\sigma_{\rm{noise}}^2 \in \{1, 2, 4, 8, 16, 32, 64\}$.

The estimator performs as intended.
Consider first the rightmost plot in Figure~\ref{fig:eigen}, where $n$ is the largest.
When the SNR is $1$, the intrinsic dimension is $5$;
in this case, we detect intrinsic dimension $5$ for $s=1,\dots,5$, and the corresponding line is roughly the constant function $s\mapsto 5$.
When the SNR is high, the mass of the design points is mostly concentrated on an $s$-dimensional subspace so that the intrinsic dimension is $s$;
the resulting estimates are then close to the identity $s\mapsto s$.
Between these extreme cases it is not clear exactly what the intrinsic dimension is.
However, the lines corresponding to the estimates behave roughly monotonically.
This means that the estimator is correctly picking up on the fact that the mass of the distribution is concentrating on a lower dimensional subspace, and gradually changing to reflect this.
For smaller sample sizes, as per the discussion from Section~\ref{sec:simulations:scaling}, the rightmost points in the plot are not reliable estimates.

\subsection{Comparison with Other Estimators}\label{sec:simulations:comparison}

We compare our estimator with competing approaches from the literature.
To do this we repeat the numerical simulations of Section 4.2 of~\citep{kleindessner2015dimensionality}, which is conceptually close to our estimator.
Our results can then be compared directly with the results from their Table 1.
Note that we base our estimates on two (symmetric) adjacency matrices $\bm A_\epsilon$ and $\bm A_{2\epsilon}$, while the estimators from~\citep{kleindessner2015dimensionality} are based on a (directed) $k$-nearest neighbour graph, so that both approaches require the same number of measurements.
We consider twelve data sets;
seven consist of simulated data, and five of real data.
Table~\ref{table:simulations} contains the results.

{\renewcommand{\arraystretch}{1.3}
\begin{table}[!tb]
\centering
\begin{tabular}{cccclc}
\hline
   	& n    	& d  	& D& \multicolumn{1}{c}{Data set} 			      & $\tilde d_n$ 	\\ \hline\hline
1	  & 1000 	& 1  	& 3        		    & Uniform on a helix      & 1.25 	(0.16)	\\
2  	& 1000 	& 2  	& 3       	   		& Swiss roll              & 2.44 	(0.59)	\\
3  	& 1000 	& 5  	& 5          			& Independent Gaussian	  & 5.26 	(1.15)	\\
4  	& 1000 	& 7  	& 8          			& Uniform on a sphere   	& 6.66 	(0.16)	\\
5  	& 5000 	& 7  	& 8        		   	& Uniform on a sphere     & 7.10 	(0.08)	\\
6  	& 1000 	& 12  & 12       			  & Uniform on $[0,1]^{12}$	& 8.70 	(0.61)	\\
7  	& 5000 	& 12  & 12       			  & Uniform on $[0,1]^{12}$	& 9.65 	(0.35)	\\ \hline
8  	& 698 	& -- 	& $64\times64$   	& Isomap faces            & 4.22	(0.70)	\\
9  	& 481  	& -- 	& $512\times480$ 	& Hands                   & 2.14 	(0.35)	\\
10 	& 7141 	& -- 	& $28\times28$  	& MNIST ``3''             & 15.62 (0.11)	\\
11 	& 6824 	& -- 	& $28\times28$   	& MNIST ``4''             & 15.65 (0.16)	\\
12 	& 6313 	& -- 	& $28\times28$   	& MNIST ``5''             & 15.45 (0.07)	\\ \hline
\end{tabular}
\caption{\footnotesize
Numerical experiments from Section 4.2 of~\citep{kleindessner2015dimensionality} for different simulated and real data sets.
For each data set we indicate the sample size, intrinsic dimension $d$, and ambient dimension $D$.
The first seven data sets are simulated, while the last five are real.
The intrinsic dimension of the real data sets is unknown.
}\label{table:simulations}
\end{table}
}

The simulated data sets are self explanatory.
The \emph{Isomap faces} data set\footnote{\url{http://isomap.stanford.edu/datasets.html}} contains 698 images ($D=64\times64$ pixels) of a rendered face of a sculpture taken from different angles, under different lighting conditions.
The \emph{Hands} data set\footnote{\url{http://vasc.ri.cmu.edu//idb/html/motion/hand/index.html}} contains 481 frames ($D=512\times480$ pixels) from a video of a hand holding a rice bowl and revolving it while moving from right to left.
The \emph{MNIST} data sets\footnote{\url{http://yann.lecun.com/exdb/mnist/}} contain $7141$, $6824$, and $6313$ images ($D=28\times28$ pixels) of handwritten digits ``3'', ``4'', and ``5'', respectively.

About the choice of $\epsilon$ for the experiments.
For every synthetic data set we set $\epsilon = 4/(\log n)^{1/2}$ as before.
For the real data sets this turned out to be inappropriate since the observations are on completely different scales.
We scaled $\epsilon$ up so that the resulting adjacency matrices $\bm A_\epsilon$ and $\bm A_{2\epsilon}$ were neither complete graphs nor empty graphs, since this would lead to trivial estimates.
This turn out to give us $\epsilon$ equal to $14$, $17$, $1600$, $1600$, and $1600$, for data sets $8$--$12$, respectively.
In practice this can be achieved based on some notion of the scale of the observations, or based on a preliminary analysis, simulated data sets, or subsamples.
In all experiments we set $m_n = 2 \log n$.\par

Our estimator compares quite well with the competing approaches, particularly as far as recuperating the integer dimension is concerned;
the results are perhaps closer to the ones obtained by the estimator of~\citep{levina2004maximum}, but contrary to their estimator we do not require any knowledge about how the distance from a given point to its $k$-nearest-neighbour scales, or any distance data, or for any parameters to be set in the estimator.
Also, the computational complexity of our estimator scales like $O(n\log n)$, considerably smaller than the typical $O(n^2)$.
Furthermore, the results in each row of Table~\ref{table:simulations} were obtained from a single data set, without repeated sampling.
Because of this we can provide standard deviations for our estimates without a need to resample the data, including for the real data sets.
These standard deviations are much more realistic than the ones associated with competing approaches.
We suspect that this is because for those it is difficult to properly balance the variance and the squared bias of the estimator, resulting in estimates that are overly concentrated around their biased means.

For the simulated data sets the true dimension is recuperated with good accuracy, with the exception of the sixth and seventh data set where the intrinsic dimension is relatively high.
Indeed, as discussed in Section~\ref{sec:simulations:scaling}, in order to recuperate the intrinsic dimension consistently, the sample size should be quite large compared to the intrinsic dimension, since the minimax rates for the problem are logarithmic in $n$.
This also comes from the fact that the support of the data set is rather unstructured, unlike for example data set $2$, $4$, and $5$, or even $10$, $11$, and $12$.

Less can be said about the results for the real data sets since the true intrinsic dimension is unknown.
However, our results are comparable to the ones obtained by competing approaches.
In particular, in all cases the intrinsic dimension is substantially smaller than the ambient dimension.
For the \emph{Isomap faces} data set we estimate the intrinsic dimension as $4$;
although the statue is 3-dimensional, the different lighting conditions may explain the fact that we detect an extra dimension in the data set.
For the \emph{Hands} data set we estimate the dimension as $2.14$.
One would probably expect the dimension to be $3$, but given the symmetries in the hand and bowl, and that the images actually make up a smooth animation may explain the lower estimate.
As for the \emph{MNIST} data set, it seems reasonable that the estimates are not too different for the three digits.
Also, if one were to parametrise the digits in terms of lengths, relative angles, and curvatures of the line segments, the estimate seems rather natural.

\section{Discussion}\label{sec:discussion}

In this paper we propose a method to estimate the intrinsic dimension of high-dimensional data sets.
The approach combines the notion of correlation dimension with the doubling property of the Lebesgue measure to provide a computationally tractable estimator for data sets with (potentially) scale-dependent dimension.
The approach does not require any parameters to be chosen, other than the scale at which one would like to estimate the dimension.
This is particularly useful for data that live on manifolds whose dimension may be different at different scales, data sets corrupted with noise, or whenever not much is known about the distribution of the data.
We compute the estimator's asymptotic distribution and rate of convergence.
The rate that we obtain matches the logarithmic minimax rate for the (easier) problem where one has access to the observations -- not just whether each pair of observations is close or not -- and is therefore optimal.
The estimator can be quickly evaluated in $O(n\log n)$ steps which is also an advantage over competing approaches, whose execution time typically scales like $O(n^2)$.
Also in terms of storage there are advantages because the adjacency matrix that we base our estimator on will typically be sparse (since the underlying graph is embedded in a Euclidean space).
Our results provide important information for algorithms commonly used to perform dimensionality reduction, learn manifolds, compress information, do statistical adaptation, and design efficient algorithms.

Distance-based estimators usually require rather (distribution specific) knowledge since one needs to know quite precisely how distances between perturbed observations scale.
These also usually require certain bandwidth parameters to be defined without an automatic, or data driven way of picking them.
Rather than assuming that we have access to the observations (or distances between them), we simply assume that we observe a graph encoding whether observations are close or not at the scale we are interested in.
This is particularly relevant when dealing with large data sets.
Modelling the resulting graph as a random connection model allows us to provide bounds on the probability of recuperating the correct intrinsic dimension of the data set, under a mild identifiability condition.

Our numerical experiments show that the intrinsic dimension can be well recuperated even without access to any distance information between the observations.
Furthermore, our estimator properly picks up on the uncertainty of the estimate (the standard deviation of the estimator can be estimated without need for resampling), which can be used to avoid the estimator to be overly concentrated around its biased mean.
Distance-based estimators tend to be much more costly, computationally.
The estimator is parameter-free, but it can easily be improved by using any knowledge one may have about the distribution of the data.
This is done by incorporating this knowledge into the choice of the function $g_\epsilon$ that features in the definition of the estimator.

Similar distance-free estimators such as those based on $k$-nearest-neighbours, seem to somewhat underperform in comparison.
Although based on a similar idea, it seems like the number of $k$-nearest-neighbours to a fixed observations does not scale in a simple way with the dimension, making it more difficult to recuperate the dimension accurately from such kind of information alone.
It is also not clear how to control exactly at which scale the dimension is being recuperated by choice of $k$.
This suggests that in data sets with different dimensions at different scales, or data sets corrupted with noise, those estimators may return some form of ``average dimension'' across scales.

Another point to make is that one can also use our work for estimation of local dimension; cf.~\citep{amsaleg2015estimating}.
These are cases where the dimension of the underlying manifold of interest is not equal everywhere, and the space is instead some heterogeneous manifold.
Our estimator is based on looking at just $m_n$ vertices in the underlying neighbourhood graph.
If we focus on vertices corresponding to the region of the manifold where we would like to estimate the local dimension (rather than picking these $m_n$ vertices arbitrarily), then our estimator promptly delivers an estimate of this local dimension.

% Acknowledgements should go at the end, before appendices and references

\paragraph{Acknowledgements}
We would like to thank the Action Editor Francois Caron and the two anonymous Referees for pointing out inconsistencies in the notation of our initial submission, for suggesting references to other closely related work, and for their useful comments and suggestions.


\appendix
\section*{Appendix}
\label{app:proofs}

In this appendix we collect the proofs to our results.

\paragraph{Proof of Theorem~\ref{theo:rate_p1}:}
By the symmetry of $\bm A_\epsilon$ it suffices to control the sum
\[
\sum_{i=1}^{m_n}\sum_{j=i+1}^n \big\{A_{\epsilon, i, j}-p_{\epsilon,1}\big\} =
\sum_{i=1}^{m_n}\sum_{j=i+1}^{m_n} \big\{A_{\epsilon, i, j} - p_{\epsilon,1}\big\} +
\sum_{i=1}^{m_n}\sum_{j=m_n+1}^n \big\{A_{\epsilon, i, j} - p_{\epsilon,1}\big\} =
(A) + (B).
\]
The proof now proceeds in two different ways:
if $m_n = n$ then $(B)=0$ and use the martingale central limit theorem to show that a properly rescaled version of $(A)$ is asymptotically standard Gaussian\footnote{The case where $O(n)\le m_n < n$ is controlled in the same way.};
if $m_n=o(n)$ then $(B)$ dominates $(A)$, so we show instead that $(B)$ is asymptotically Gaussian.\par

Let $m_n = n$.
By the symmetry of $\bm A_\epsilon$, and since $A_{\epsilon, i, i}=0$, $(A)$ can be rewritten as
\[
\sum_{i=2}^n\left[ \sum_{j=1}^{i-1}\big\{A_{\epsilon, i, j}-p_\epsilon(X_j)\big\} +
(n-i)\big\{p_\epsilon(X_i)-p_{\epsilon,1}\big\} \right] + (n-1)\big\{p_\epsilon(X_1)-p_{\epsilon,1}\big\}.
\]
(Note that we are just adding and subtracting $p_\epsilon(X_j)$ inside the double sum and reordering terms.)
Denote the term in square brackets as $\eta_{n, i}$.
The two terms in the previous display are uncorrelated;
the variance of the second term is $(n-1)^2(p_{\epsilon,2}-p_{\epsilon,1}^2)$.
Further, $\eta_{n, i}$ has expectation $0$, and is measurable with respect to $\mathcal{F}_{n, i}=\mathcal{F}_i=\sigma(X_1,\dots,X_i)$.
Since $j<i$,
\[
\mathbb{E}\big[\eta_{n, i}\mid\mathcal{F}_{i-1}\big] =
\sum_{j=1}^{i-1}\big\{\mathbb{E}\big[A_{\epsilon, i, j}\mid X_j\big]-p_\epsilon(X_j)\big\} + (n-i)\big\{\mathbb{E}p_\epsilon(X_i)-p_{\epsilon,1}\big\} = 0, \quad i=2,\dots,n,
\]
so that $\eta_{n, i}$ is a martingale increment.
This means that if we define $S_{n, k}=\sum_{i=2}^k\eta_{n, i}$, then $\{S_{n, k},\mathcal{F}_{n, k},2\le k\le n, n\ge 2\}$ is a martingale array.

We show that a scaled version of $S_{n, n}$ is asymptotically standard Gaussian, as $n\to\infty$.
With the convention that $p_\epsilon(X_i,X_i)=p_\epsilon(X_i)$, by expanding the square and taking conditional expectations term-wise it holds that
\begin{equation}\label{eq:conditional_squared_increments}
\begin{aligned}&
\mathbb{E}\big[\eta_{n, i}^2\mid \mathcal{F}_{i-1}\big] =
\sum_{j_1=1}^{i-1}\sum_{j_2=1}^{i-1}\big\{p_\epsilon(X_{j_1},X_{j_2})-p_\epsilon(X_{j_1})p_\epsilon(X_{j_2})\big\} +\\ & \qquad\qquad+
2(n-i)\sum_{j=1}^{i-1}\big\{\mathbb{E}\big[A_{\epsilon, i, j}\,p_\epsilon(X_i)\mid X_j\big]-p_{\epsilon,1}\,p_\epsilon(X_j)\big\} +
(n-i)^2(p_{\epsilon,2}-p_{\epsilon,1}^2),
\end{aligned}
\end{equation}
where we use the definitions of $p_\epsilon(X)$ and $p_\epsilon(X,Y)$, the fact that $p_\epsilon(X_j)\in \mathcal{F}_{n, i-1}$, for $j\le i-1$, and that $X_i$ -- or indeed $p_\epsilon(X_i)$ -- is independent of $\mathcal{F}_{n, i-1}$, $i=2,\dots, n$.

The expectation of $\eta_{n, i}^2$ is obtained by taking expectation of the previous display and noting that for any $j_1,j_2,j\neq i$,
\begin{align*}
\mathbb{E}\big\{p_\epsilon(X_{j_1},X_{j_2})-p_\epsilon(X_{j_1})p_\epsilon(X_{j_2})\big\} &=
\left\{p_{\epsilon,1} - \mathbb{V}P_\epsilon(X)\right\} 1_{\{j_1=j_2\}} + p_{\epsilon,2}1_{\{j_1\neq j_2\}} - p_{\epsilon,1}^2,\\
\mathbb{E}\big\{\mathbb{E}\big[A_{\epsilon, i, j}\,p_\epsilon(X_i)\mid X_j\big]-p_{\epsilon,1}\,p_\epsilon(X_j)\big\} &=
p_{\epsilon,2}-p_{\epsilon,1}^2.
\end{align*}
Note that $\sum_{i=2}^n(i-1)$ is $n^2\{1/2+o(1)\}$, and the sums $\sum_{i=2}^n(n-i)^2$, $2\sum_{i=2}^n(n-i)(i-1)$, and $\sum_{i=2}^n(i-1)(i-2)$ are all $n^3\{1/3+o(1)\}$.
We assume\footnote{
This assumption is weaker than the assumption $n(p_{\epsilon,2} - p_{\epsilon,1}^2)^2\to\infty$ which will be imposed later.
} that $p_{\epsilon,1}-p_{\epsilon,2}=o\{n(p_{\epsilon,2}-p_{\epsilon,1}^2)\}$, as $n\to\infty$ so that the variance of $S_{n, n}$ becomes
\[
\mathbb{V}S_{n, n} = \sum_{i=2}^n \mathbb{E}\eta_{n, i}^2 =
\left\{\frac{n^2}2(p_{\epsilon,1}-p_{\epsilon,1}^2) + n^3(p_{\epsilon,2}-p_{\epsilon,1}^2)\right\}\{1+o(1)\} =
n^3(p_{\epsilon,2}-p_{\epsilon,1}^2)\{1+o(1)\}.
\]
We therefore define $Z_{n, i} = S_{n, i}/\{\mathbb{V}S_{n, n}\}^{1/2}$, with increments $\xi_{n, i}=\eta_{n, i}/\{\mathbb{V}S_{n, n}\}^{1/2}$.

Based on the preceding, $\{Z_{n, i},\mathcal{F}_{n, i},2\le i\le n, n\ge 2\}$ is a (zero-mean, unit variance) martingale array.
Since $\mathcal{F}_{n, i}=\mathcal{F}_{i}$, the $\sigma$-fields satisfy $\mathcal{F}_{n, i} \subseteq \mathcal{F}_{n+1,i}$, $2\le i\le n$, $n\ge 2$, so that they are nested.
We check the conditions of Corollary 3.1 of~\citep{hall2014martingale}:
\begin{equation}\label{eq:conditions_MCLT}
\sum_{i=2}^n \mathbb{E}\big[\xi_{n, i}^2 \indic_{\{|\xi_{n, i}|>\delta\}} \mid \mathcal{F}_{i-1}\big] \stackrel{P}{\longrightarrow}0, \quad \delta>0,
\qquad\hbox{and}\qquad
\sum_{i=2}^n \mathbb{E}\big[\xi_{n, i}^2 \mid \mathcal{F}_{i-1}\big] \stackrel{P}{\longrightarrow}1.
\end{equation}

To check the first condition we use $\mathbb{E}\big[|Z|^2\indic_{\{|Z|>\delta\}}\big] \le \mathbb{E}Z^4/\delta^2$ (by the Cauchy-Schwarz inequality, and Markov's inequality), and $(a+b)^4\le 2^3(a^4+b^4)$, $a, b\in\mathbb{R}$ (by Young's inequality).
Therefore, checking~\eqref{eq:conditions_MCLT} reduces to showing that as $n\to\infty$,
\[
\sum_{i=2}^n\frac{\mathbb{E}\big[ \big(\sum_{j=1}^{i-1}\big\{A_{\epsilon, i, j}-p_\epsilon(X_j)\big\}\big)^4 \mid\mathcal{F}_{i-1}\big]}{n^6(p_{\epsilon,2}-p_{\epsilon,1}^2)^2}\stackrel{P}{\longrightarrow}0, \qquad
\sum_{i=2}^n\frac{(n-i)^4\mathbb{E}\big[ \big(p_\epsilon(X_i)-p_{\epsilon,1}\big)^4\big]}{n^6(p_{\epsilon,2}-p_{\epsilon,1}^2)^2}\longrightarrow0.
\]
If $\epsilon$ is fixed, then $p_{\epsilon,1}$ and $p_{\epsilon,2}$ are fixed, the numerators are of order $n^5$ and the conditions are met.
Assume therefore that $\epsilon\to0$ so that $p_\epsilon(x)\to 0$.
%By Markov's inequality, we can drop the conditioning on the left and just show that the expectation of the ratio converges to zero.
Since the summands are positive, both conditions follow if we assume that $n(p_{\epsilon,2}-p_{\epsilon,1}^2)^2\to\infty$.

To show the second condition in~\eqref{eq:conditions_MCLT} holds, it suffices to show that the sum over $i=2,\dots, n$ of each of the three terms in~\eqref{eq:conditional_squared_increments} divided by $n^3(p_{\epsilon,2}-p_{\epsilon,1}^2)$ converges to $1/3$, in probability;
this is obvious for the third term, for the second term it follows easily from Chebyshev's inequality (since the terms in the sum are independent), so that only the convergence of the first term requires justification.

By Chebyshev's inequality and symmetry, it suffices to show that
\[
\frac{\mathbb{E}\left(\sum_{i=2}^n\sum_{j_1=1}^{i-1}\sum_{j_2=1}^{i-1}\big\{p_\epsilon(X_{j_1},X_{j_2})-p_\epsilon(X_{j_1})p_\epsilon(X_{j_2}) -(p_{\epsilon,2}-p_{\epsilon,1}^2)\big\}\right)^2}{n^6(p_{\epsilon,2}-p_{\epsilon,1}^2)^2}
\]
converges to zero.
By Jensen's inequality, we can upper bound the previous display by
\[
(n-1)\sum_{i=2}^n\frac{\mathbb{E}\left(\sum_{j_1=1}^{i-1}\sum_{j_2=1}^{i-1}\big\{p_\epsilon(X_{j_1},X_{j_2})-p_\epsilon(X_{j_1})p_\epsilon(X_{j_2}) -(p_{\epsilon,2}-p_{\epsilon,1}^2)\big\}\right)^2}{n^6(p_{\epsilon,2}-p_{\epsilon,1}^2)^2} \lesssim \frac1{n(p_{\epsilon,2}-p_{\epsilon,1}^2)^2}.
\]
The last equality follows from the fact that of the $(i-1)^4$ involved in the square above, $(i-1)^2(i-3)^2$ of them have zero mean; the remaining terms are $O(i^3)$.
The previous display is $o(1)$ since we assume $n(p_{\epsilon,2}-p_{\epsilon,1}^2)^2\to\infty$.\par

To complete the proof, we consider the case $m_n=o(n)$.
Adding and subtracting $p_\epsilon(X_i)$ inside the double sum, and then interchanging the two summations we can rewrite $(B)$ as
\[
\sum_{j=m_n+1}^n\left[\sum_{i=1}^{m_n}\big\{A_{\epsilon, i, j}-p_\epsilon(X_i)\big\}\right] +
(n-m_n)\sum_{i=1}^{m_n}\big\{p_\epsilon(X_i)-p_{\epsilon,1}\big\} = (C) + (D).
\]
Since the terms in $(D)$ are independent, it is easy to see that the variance of $(D)$ is $m_n(n-m_n)^2(p_{\epsilon,2}-p_{\epsilon,1}^2)$, so that it dominates the terms in $(A)$, and is asymptotically Gaussian.
It remains to show that the variance of $(C)$ is dominated by that of $(D)$.

Define the sum in square brackets in the previous display as $\zeta_j$.
These terms are uncorrelated:
for indices $j_1\neq j_2$, with $j_1,j_2>i_1,i_2$,
\begin{align*}
\mathbb{E}\zeta_{j_1}\zeta_{j_2} =&
\sum_{i=1}^{m_n}\mathbb{E}\big\{A_{\epsilon, i, j_1}-p(X_i)\big\}\big\{A_{\epsilon, i, j_2}-p(X_i)\big\}\\ & +
\sum_{i_1=1}^{m_n}\sum_{\substack{i_2=1\\i_2\neq i_1}}^{m_n}
\mathbb{E}\big\{A_{\epsilon,i_1,j_1}-p_\epsilon(X_{i_1})\big\}\big\{A_{\epsilon,i_2,j_2}-p_\epsilon(X_{i_2})\big\},
\end{align*}
which is zero.
From this we conclude that the variance of $(C)$ is $O\{nm_n^2(p_{\epsilon,2}-p_{\epsilon,1}^2)\}$, and is therefore dominated by that of $(D)$.
This concludes the proof.\par

\paragraph{Proof of Theorem~\ref{theo:consistency_hat_d}:}
We apply the delta method.
For this we need to know the joint distribution of $\{\hat p_{n,\epsilon,1}, \hat p_{n,2\epsilon,1}\}$.
This is established using the Cram\'er-Wold device by showing that for each $\alpha,\beta\in\mathbb{R}$, $\alpha\, \hat p_{n,\epsilon,1} + \beta\, \hat p_{n,2\epsilon,1}$ is asymptotically Gaussian.
As in the proof of Theorem~\ref{theo:rate_p1} we control
\[
\sum_{i=1}^{m_n}\sum_{j=i+1}^n \big\{\alpha A_{\epsilon, i, j}+\beta A_{2\epsilon, i, j}-\alpha p_{\epsilon,1} - \beta p_{2\epsilon,1}\big\} = \sum_{i=1}^{m_n}\sum_{j=i+1}^n \big\{A_{\epsilon, i, j}^*- p_{\epsilon,1}^*\big\}.
\]
Since $A_{\epsilon, i, j}^*$ is measurable with respect to $X_i$ and $X_j$, and bounded, and since the previous display has mean zero, we can follow exactly the same steps as in Theorem~\ref{theo:rate_p1} to show that the previous display is asymptotically Gaussian.

To fully specify the asymptotic distribution of $\{\hat p_{n,\epsilon,1}, \hat p_{n,2\epsilon,1}\}$ it remains to compute the covariance of the estimates $\hat p_{n,\epsilon,1}$ and $\hat p_{n,2\epsilon,1}$.
Define $p_{\epsilon_1,\epsilon_2,2}=\mathbb{P}\{r(X,Z)\le\epsilon_1,r(Z,Y)\le\epsilon_2\}$,
and assume that $p_{\epsilon,1}-p_{\epsilon,1}\cdot p_{2\epsilon,1} = o\big[n\{p_{\epsilon,2\epsilon,2}-p_{\epsilon,1}\cdot p_{2\epsilon,1}\} \big]$.
Simple computations then give
\[
\mathbb{V}\big\{\hat p_{n,\epsilon,1},\, \hat p_{n,2\epsilon,1}\big\} =
\frac{(n+3m_n)\{p_{\epsilon,2\epsilon,2}-p_{\epsilon,1}p_{2\epsilon,1}\}}{n\cdot m_n}\{1+o(1)\}.
\]
Write
$A_\epsilon = p_{\epsilon,2}-p_{\epsilon,1}^2$,
$B_\epsilon = p_{2\epsilon,2}-p_{2\epsilon,1}^2$, and
$C_\epsilon = (n+3m_n)\{p_{\epsilon,2\epsilon,2} - p_{\epsilon,1}\cdot p_{2\epsilon,1}\}/n$.
Conclude that under the assumptions of Theorem~\ref{theo:rate_p1} (verified also with $\epsilon$ replaced with $2\epsilon$),w
\[
m_n^{1/2}\left\{
\begin{bmatrix}\hat p_{n,\epsilon,1}\\ \hat p_{n,2\epsilon,1}\end{bmatrix} - \begin{bmatrix}p_{\epsilon,1}\\ p_{2\epsilon,1}\end{bmatrix}\right\}
\stackrel{d}{\longrightarrow}
N\left\{
\begin{bmatrix}0\\ 0\end{bmatrix}, \;
\begin{bmatrix}A_\epsilon & C_\epsilon\\ C_\epsilon & B_\epsilon\end{bmatrix}
\right\}.
\]
In the general case of the implicit estimates in~\eqref{def:estimators_d_implicit}, consider $d\mapsto g_\epsilon(d)$;
let $g_\epsilon^{-1}$ represent the inverse of $g_\epsilon$ (which exists, at least in a neighbourhood of $d$).
The asymptotic distribution of the estimator $\hat d_n$ is obtained by applying the delta method to the previous display using the function $(\alpha,\beta)\mapsto g_\epsilon^{-1}(\beta/\alpha)$, whose gradient is $\{\partial \log g_\epsilon(d)/\partial d\}^{-1} [-1/\alpha, 1/\beta]^T$.
This delivers the asymptotic distribution of
\[
g_\epsilon^{-1}(\hat p_{n,\epsilon,1}/\hat p_{n,2\epsilon,1}) -
g_\epsilon^{-1}(p_{\epsilon,1}/p_{2\epsilon,1}) =
\hat d_n - d + d - g_\epsilon^{-1}(p_{\epsilon,1}/p_{2\epsilon,1}) =
\hat d_n - d + o(m_n^{-1/2}),
\]
where the last equality follows from the bias condition~\eqref{eq:bias}.
We conclude that
\[
m_n^{1/2}\left\{ \hat d_n - d \right\} \stackrel{d}{\longrightarrow}
N\left\{0,\, \left\{\frac{\partial\log g_\epsilon(d)}{\partial d}\right\}^{-2}\left[
\frac{A_\epsilon}{p_{\epsilon,1}^2} +
\frac{B_\epsilon}{p_{2\epsilon,1}^2}
- \frac{2C_\epsilon}{p_{\epsilon,1}\, p_{2\epsilon,1}}\right] \right\}.
\]
Replacing $A_\epsilon$, $B_\epsilon$, and $C_\epsilon$, the component of the variance in square brackets is
\begin{equation}\label{eq:variance_d_general}
6\frac{m_n}n +
\frac{p_{\epsilon,1}^2p_{2\epsilon,2}+p_{2\epsilon,1}^2p_{\epsilon,2}}{p_{\epsilon,1}^2p_{2\epsilon,1}^2}-
2\left(1+3\frac{m_n}n\right)\frac{p_{\epsilon,2\epsilon,2}}{p_{\epsilon,1}p_{2\epsilon,1}}.
\end{equation}

For the explicit estimator from~\eqref{def:estimators_d_explicit}, the delta method is used with the function $(\alpha,\beta)\mapsto \log(\beta/\alpha)/\log(2) =g_\epsilon^{-1}(\beta/\alpha)$, such that the scaling in the variance becomes $\log(2)^{-2}$.\par

\vskip 0.2in
\bibliography{refs}

\end{document}